\documentclass[10pt,twocolumn,letterpaper]{article}

\usepackage{iccv}
\usepackage{times}
\usepackage{epsfig}
\usepackage{graphicx}
\usepackage{amsmath}
\usepackage{amssymb}

\usepackage{booktabs}
\usepackage{bm}
\usepackage{makecell}
\usepackage{enumitem}
\usepackage{multirow}
\makeatletter
\@namedef{ver@everyshi.sty}{}
\makeatother
\usepackage{booktabs}
\usepackage{multirow}
\usepackage[normalem]{ulem}
\usepackage{ifsym}

\usepackage{tikz}
\usepackage{comment}

\usepackage[accsupp]{axessibility}  % Improves PDF readability for those with disabilities.

\usepackage{color}
\usepackage{contour}
\usepackage{ulem}
\usepackage{textcomp}

\usepackage{nopageno}

\contourlength{0.5pt}

\newcommand{\culine}[1]{%
  \uline{\phantom{#1}}%
  \llap{\contour{white}{#1}}%
}

\definecolor{todo}{rgb}{1.0, 0., 0.}

\definecolor{reff}{rgb}{1.0,0.0,0.0}
\newcommand{\reff}[1]{\textcolor{reff}{{{#1}}}}

\newcommand\blfootnote[1]{%
\begingroup
\renewcommand\thefootnote{}\footnote{#1}%
\addtocounter{footnote}{-1}%
\endgroup
}

\usepackage[pagebackref=true,breaklinks=true,letterpaper=true,colorlinks,bookmarks=false]{hyperref}

\iccvfinalcopy % *** Uncomment this line for the final submission

\def\iccvPaperID{9915} % *** Enter the ICCV Paper ID here

\ificcvfinal\fi

\begin{document}

%%%%%%%%% TITLE
\def\pname{U3HS}

%Holistic Segmentation\\
\title{Segmenting Known Objects and Unseen Unknowns without Prior Knowledge}

\author{Stefano Gasperini$^{1,2,\circ}$ \quad 
Alvaro Marcos-Ramiro$^{2}$ \quad
Michael Schmidt$^{2}$ \quad \\
Nassir Navab$^{1}$ \quad 
Benjamin Busam$^{1}$ \quad 
Federico Tombari$^{1,3}$\\\\
$^1$ Technical University of Munich \quad $^2$ BMW Group \quad $^3$ Google}
%\institute{Technical University of Munich (TUM), Munich, Germany
%\and
%BMW Group, Munich, Germany
%\and
%Google, Zurich, Switzerland}
%\end{comment}
%******************
\maketitle

\blfootnote{$^{\circ}$ This work was conducted while working at BMW Group.}
\blfootnote{Contact author: Stefano Gasperini (\textit{stefano.gasperini@tum.de}).}

%%%%%%%%% ABSTRACT

\begin{abstract}
Panoptic segmentation methods assign a known class to each pixel given in input. Even for state-of-the-art approaches, this inevitably enforces decisions that systematically lead to wrong predictions for objects outside the training categories. However, robustness against out-of-distribution samples and corner cases is crucial in safety-critical settings to avoid dangerous consequences. Since real-world datasets cannot contain enough data points to adequately sample the long tail of the underlying distribution, models must be able to deal with unseen and unknown scenarios as well. Previous methods targeted this by re-identifying already-seen unlabeled objects. In this work, we propose the necessary step to extend segmentation with a new setting which we term holistic segmentation. Holistic segmentation aims to identify and separate objects of unseen, unknown categories into instances without any prior knowledge about them while performing panoptic segmentation of known classes. We tackle this new problem with \pname, which finds unknowns as highly uncertain regions and clusters their corresponding instance-aware embeddings into individual objects. By doing so, for the first time in panoptic segmentation with unknown objects, our \pname\ is trained without unknown categories, reducing assumptions and leaving the settings as unconstrained as in real-life scenarios. Extensive experiments on public data from MS COCO, Cityscapes, and Lost\&Found demonstrate the effectiveness of \pname\ for this new, challenging, and assumptions-free setting called holistic segmentation. Project page: \href{https://holisticseg.github.io/}{https://holisticseg.github.io}.
\end{abstract}

%%%%%%%%% BODY TEXT

\section{Introduction} \label{sec:intro}
% recent focus on robustness and out-of-domain data
Since neural networks have achieved unprecedented performance in perception tasks (e.g., object detection and semantic segmentation), there has been a growing interest in ensuring their safe deployment, especially important for safety-critical scenarios, such as autonomous driving and robotics~\cite{gawlikowski2021survey}. Recently, several works have been proposed to improve robustness and generalization by addressing corner cases and out-of-distribution data~\cite{cen2021openset3ddetection,wang2022generalizing,gasperini2021r4dyn}, via domain adaptation~\cite{wu2021dannet}, adversarial augmentations~\cite{lehner20223d}, simulations~\cite{beery2020synthetic}, and uncertainty estimation~\cite{postels2020hidden}.

\begin{figure}[t]
\begin{center}
\includegraphics[width=1.00\linewidth]{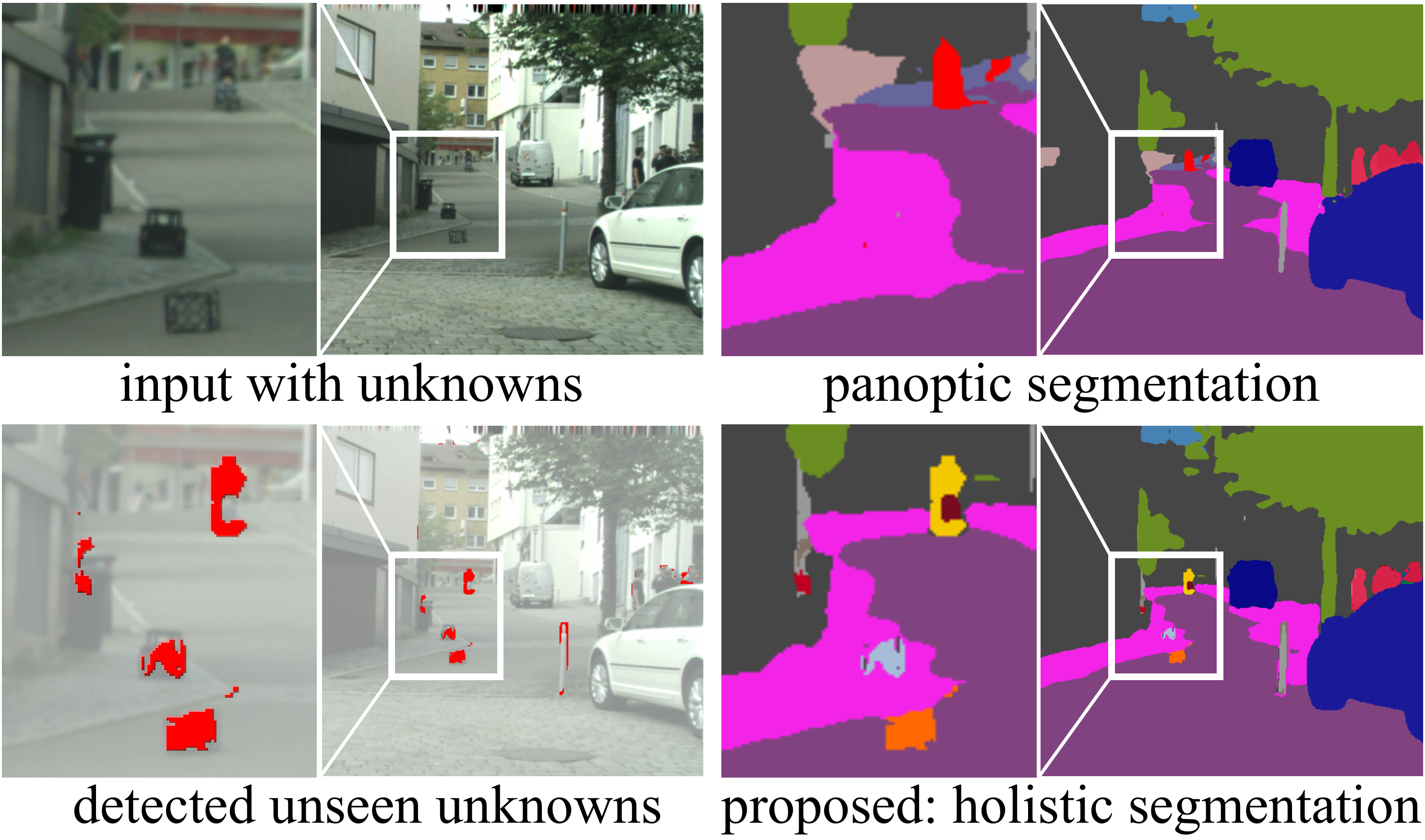}
\vspace{-0.7cm}
\end{center}
   \caption{
   State-of-the-art panoptic segmentation methods~\cite{cheng2020panopticdeeplab} cannot deal with unseen classes from~\cite{pinggera2016lost} (top right). Instead, our \pname\ addresses the proposed holistic segmentation setting. \pname\ finds unseen unknowns (bottom left) and separates them into instances (bottom right) without prior knowledge about unknowns.
   }
\label{fig:teaser}
\vspace{-0.3cm}
\end{figure}

% datasets and corner cases
Due to the difficulty of collecting corner cases from the long tail of the underlying data distribution, current datasets cannot fully represent the diversity of the world, leaving its vast majority as difficult out-of-distribution (OOD) samples~\cite{hendrycks2021natural,chan2021segmentmeifyoucan}. In safety-critical applications, considering them during development and deployment is of utmost importance~\cite{blum2021fishyscapes,lehner20223d}, or they could cause severe damage.

% the role of uncertainty estimation in safety critical scenarios (e.g. autonomous driving)
Furthermore, since the powerful and popular \textit{softmax} highly promotes the probability of the highest \textit{logit}, state-of-the-art methods tend to be overly confident even on wrong predictions~\cite{sensoy2018dpn,gawlikowski2021survey}. In safety-critical settings,
reliable confidence together with interpretability techniques \cite{fong2019understanding,zhang2021fine} increases trust for downstream tasks~\cite{gawlikowski2021survey}, e.g., trajectory prediction and path planning. Towards this end, estimating the uncertainty of a model's output is commonly considered a key enabler for its safe applicability~\cite{kendall2017uncertainties,gawlikowski2021survey}.

% dense tasks like segmentation and the importance of detecting the unknown -> open set: what it is and why we need it
While several works addressed some of these problems for image classification~\cite{sensoy2018dpn,liu2020sngp,postels2020hidden,van2020uncertainty} and object detection~\cite{miller2019evaluating,gasperini2021certainnet}, they remain primarily unexplored for dense tasks such as semantic and panoptic segmentation~\cite{cen2021mmsp,wong2020uberlidaropenps}. Compared to object detection, the problem is more severe for dense tasks, where a model needs to provide a prediction for every input unit (e.g., each pixel). So unseen objects (i.e., of new, unseen categories) are systematically and wrongly assigned to one of the limited number of known classes (closed-set), as shown in Figure~\ref{fig:teaser}.
This has led researchers towards designing new methods that work not only with the available data distribution but also with OOD samples that are not available (open-set), thereby improving robustness against unseen scenarios~\cite{jung2021sml,liu2020sngp,cen2021mmsp,lehner20223d}.

% the few works on open set panoptic and their limitations (relying on ood at training time)
Open-set panoptic segmentation~\cite{wong2020uberlidaropenps} segments instances of unlabeled objects in addition to panoptic segmentation of known areas, i.e., the combination of semantic and instance segmentation~\cite{kirillov2019panoptic}.
Unlike OOD segmentation~\cite{jung2021sml}, segmenting unknown instances enables tracking and trajectory prediction.
Prior works~\cite{wong2020uberlidaropenps,hwang2021exemplar,xu2022openpano_twostage} tackled this problem by relying on seeing unlabeled categories during training. They learned these categories through the \textit{void} class (i.e., unlabeled) and assumed unknowns to be within ground truth \textit{void} regions at training time and inside \textit{void} predictions at test time. By doing so, unknowns are transformed into learned unlabeled instances (i.e., essentially known objects)~\cite{hwang2021exemplar}, constraining the open-set task.
Mainly intended to segment already-seen unlabeled objects~\cite{hwang2021exemplar,xu2022openpano_twostage}, current works cannot deal with the wide variability of unknowns and corner cases outside the training data.

% in this paper...
In this paper, we propose the necessary next step for panoptic segmentation to include object categories outside the training data (i.e., unseen unknowns). We term the new setting holistic segmentation. The aim is to identify and segment unseen unknowns into instances while segmenting known classes in a panoptic fashion without any external nor prior knowledge about unknowns.
%In this paper, we broaden the problem by proposing holistic segmentation: a comprehensive and challenging task to identify instances from completely unseen object categories and perform panoptic segmentation of known areas, without learning from unknowns nor using any information about them.
%An output example is shown in Figure~\ref{fig:teaser}.
Unseen categories pose new challenges compared to already-seen unlabeled ones \cite{hwang2021exemplar}, requiring new solutions. Estimating the uncertainty is a key step towards finding the knowledge boundaries of a model, leaving the problem unconstrained and reducing assumptions on the training data.
Therefore, we propose \pname:
%the first method targeting holistic segmentation. \pname\ stands for 
\culine{U}nseen \culine{U}nknowns via \culine{U}ncertainty estimation for \culine{H}olistic \culine{S}egmentation.
%By also quantifying the model uncertainty, \pname\ constitutes a step towards safe and comprehensive scene understanding. 
The main contributions of this paper can be summarized as follows:
\begin{itemize}
    \item We introduce the setting of holistic segmentation, which highlights the importance of not using prior knowledge about unknown objects (e.g., text), and leaves the setup unconstrained as in real scenarios.
    \item We tackle this new setting with \pname: the first panoptic framework to deal with unseen, unknown object categories, able to segment and separate them.
    \item We provide uncertainty measures for the output of \pname\ to further improve its safe applicability. % in safety-critical settings, such as autonomous driving.
    %Additionally, we estimate the model uncertainty, leading to a comprehensive dense output and further improving \pname\ applicability in safety-critical settings.
\end{itemize}

%%% the hassle of manually labeling the unknown
% instantiating never seen before unknowns (first)
% novel open set panoptic method, first not seeing unknowns during training!!

\section{Related Work} \label{sec:rel_work}

%\begin{itemize}
%    \item closed set panoptic segmentation
%    \item uncertainty estimation
%    \item open set object detection
%    \item open set semantic segmentation
%    \item open set panoptic segmentation
%\end{itemize}

\textbf{Closed-set panoptic segmentation}
Combining semantic and instance segmentation, panoptic segmentation~\cite{kirillov2019panoptic} distinguishes \textit{things} (countable classes) from \textit{stuff} (amorphous).
%By combining semantic and instance segmentation, Kirillov et al.~\cite{kirillov2019panoptic} proposed panoptic segmentation. They distinguished between \textit{thing} (countable) and \textit{stuff} (amorphous) classes.
%, as countable objects and amorphous regions respectively.
%Most methods are bottom-up or top-down. 
The vast majority of methods are top-down~\cite{xiong2019upsnet,sofiiuk2019adaptis,porzi2019seamless,kirillov2019panopticfpn,hou2020realtimepano,mohan2021efficientps}: two-stage exploiting box proposals and \textit{thing} masks from Mask R-CNN~\cite{he2017maskrcnn}, and filling up \textit{stuff} areas with a semantic branch.
%Examples are EfficientPS~\cite{mohan2021efficientps} and Panoptic FPN~\cite{kirillov2019panopticfpn}. 
Bottom-up are proposal-free, e.g., Panoptic-DeepLab~\cite{cheng2020panopticdeeplab}:
%and Axial-DeepLab~\cite{wang2020axialdeeplab}. 
they segment semantically and cluster instances within \textit{thing} regions~\cite{cheng2020panopticdeeplab,wang2020axialdeeplab}.
A different line of work proposed end-to-end solutions~\cite{gasperini2021panoster,wang2021maxdeeplab} where instance and semantic segments are delivered directly by treating instance segmentation as a class-agnostic classification task.
Others explored self-attention~\cite{hou2020realtimepano}, videos~\cite{woo2021videopano,chen2020naivestudent,miao2022video_pano}, scene graphs~\cite{wu2021scenegraphfusion,wu2023incremental}, multi-task learning~\cite{de2021partawarepano,goel2021quadronet}, neural fields~\cite{kundu2022panoptic_nerf}, or text descriptions~\cite{ding2022open_maskclip}.
%Others extended dense object detection using self-attention~\cite{hou2020realtimepano}, focused on time consistency for predictions on videos~\cite{woo2021videopano}, leveraged semi-supervised learning in unlabeled videos~\cite{chen2020naivestudent}, extended panoptic predictions to incremental 3D scene graphs~\cite{wu2021scenegraphfusion}, or leveraged multi-task learning with part segmentation~\cite{de2021partawarepano} or depth estimation and 2D detection~\cite{goel2021quadronet}.
Our \pname\ framework deals with unseen unknowns and extends~\cite{cheng2020panopticdeeplab} via instance-aware embeddings.

\begin{figure}[t]
\begin{center}
\includegraphics[width=1.00\linewidth]{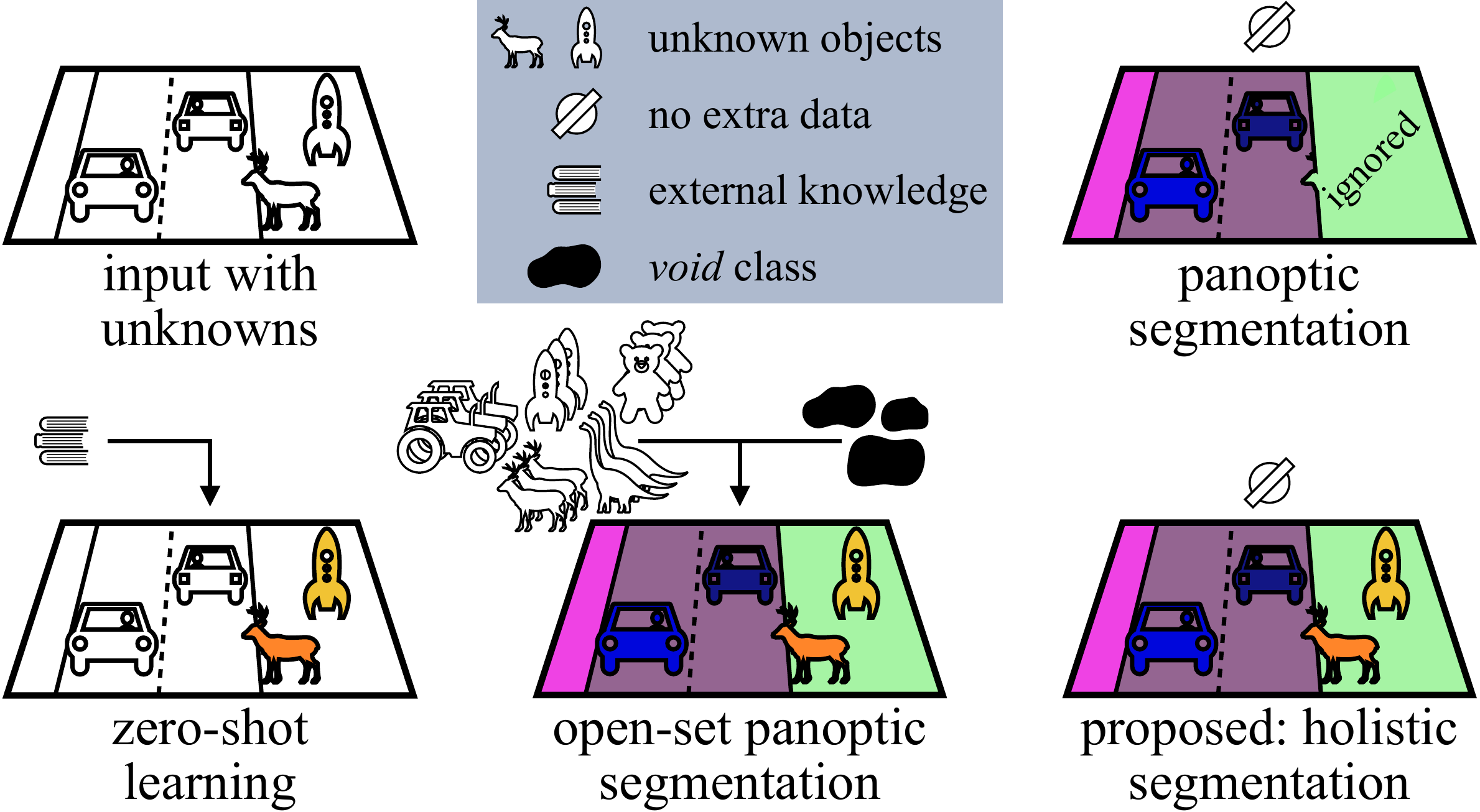}
\vspace{-0.7cm}
\end{center}
   \caption{
   Comparison between closed-set (top right)~\cite{kirillov2019panoptic} and open-set~\cite{hwang2021exemplar} panoptic segmentation, zero-shot learning~\cite{zheng2021zero_instance}, and the proposed holistic segmentation setting. While zero-shot and open-set panoptic methods commonly leverage knowledge about unknown objects, holistic segmentation does not use any priors.
   %Unlike others, holistic segmentation aims to identify instances of unknowns without prior knowledge about them.
   %, while performing panoptic segmentation on known areas.
   }
\label{fig:setting}
\vspace{-0.2cm}
\end{figure}

\textbf{Zero-shot learning} aims to predict unseen classes outside the training set~\cite{bucher2019zero_semantic,zheng2021zero_instance,yue2021counterfactual_zeros} with the help of external knowledge~\cite{chen2021knowledge_zeros_survey,geng2023benchmarking_zeros}, e.g., a language model~\cite{zhai2022lit}, used to build semantic spaces common between seen and unseen classes~\cite{yi2022hierarchical_zeros}.
While zero-shot methods detect only unseen classes at inference time, generalized zero-shot approaches also detect seen ones~\cite{chao2016generalized_zeros,pourpanah2022review_generalized_zeros}, similarly to the proposed holistic segmentation.
Also \textbf{open-vocabulary} methods are zero-shot~\cite{huynh2022openvocabulary,ghiasi2022scaling_open_voc,xu2022open_voc,ding2022open_maskclip}. They exploit language models such as CLIP~\cite{radford2021clip} to describe unknowns. CLIP has been trained on unknown classes and is treated as an oracle, as it is assumed to be able to describe every unknown, allowing open-vocabulary and zero-shot approaches to identify them. However, this implies that unknown classes are known, e.g., to CLIP.
%, which means they could be directly part of the set of known and learned classes.
Moreover, CLIP is not immune to corner cases and long tail samples~\cite{radford2021clip}. This limits the pool of objects that these methods can recognize.
As shown in Figure~\ref{fig:setting}, holistic segmentation segments unseen objects too, but unlike zero-shot and open-vocabulary, it does not use any external support (e.g., text descriptions of unknowns), such that objects of unseen categories are segmented solely by learning on known ones.
More recent than this work, SAM~\cite{kirillov2023sam} is a strong foundation model. Unlike SAM, ours does not use any prompts and outputs semantic classes.

\textbf{Uncertainty estimation}
%There are two types of uncertainty: epistemic caused by the model itself, and aleatoric due to the input~\cite{kendall2017uncertainties,gawlikowski2021survey}.
Epistemic uncertainty is caused by the model itself, while aleatoric is due to the input~\cite{kendall2017uncertainties,gawlikowski2021survey}.
OOD data typically results in high epistemic due to a knowledge gap.
%Methods can be categorized in sampling-based (e.g., MC Dropout~\cite{gal2016mcdropout}) and sampling-free~\cite{postels2019sampling}, depending on whether they aggregate multiple predictions or not. The latter better suits real-time applications and includes single deterministic approaches, which provide both prediction and uncertainty estimates with the same model~\cite{gawlikowski2021survey}. 
Single deterministic approaches are sampling-free and provide predictions and uncertainty estimates with the same model~\cite{gawlikowski2021survey}.
Among these, DUQ~\cite{van2020uncertainty} learns class representatives and compares them with input features. SNGP~\cite{liu2020sngp} improves the awareness to domain shifts via weight normalization and a Gaussian process. DPN~\cite{sensoy2018dpn} predicts the parameters of a Dirichlet distribution,
%instead of a categorical one, 
and uses a Dirichlet density function for each probability assignment and its uncertainty.
Various works estimated uncertainty for object detection~\cite{miller2018opensetdetection,miller2019evaluating,gasperini2021certainnet}, and segmentation~\cite{postels2019sampling,cen2021mmsp,sirohi2022uncertainty}, improving robustness and generalization. While most compute uncertainty only to provide it as extra output~\cite{gawlikowski2021survey,sirohi2022uncertainty}, our \pname\ can be paired with any of the above techniques to find unknown objects, which it then separates into instances for holistic segmentation.
%Exemplified with DPN~\cite{sensoy2018dpn}, which we extended and modified for this setting, 
%\pname\ supports any of the above techniques.

\textbf{Open-set perception}
Open-set tasks are similar to generalized zero-shot learning~\cite{chao2016generalized_zeros}, with the fundamental difference that here no external knowledge on the unseen classes is used~\cite{yue2021counterfactual_zeros}: inference is based only on what was learned from the training data. Uncertainty estimation helps to identify knowledge boundaries \cite{miller2018opensetdetection}.
Bayesian SSD~\cite{miller2018opensetdetection} uses Dropout sampling for open-set object detection. MLUC~\cite{cen2021openset3ddetection} tackles this for LiDAR point clouds via metric learning and unsupervised clustering.
Open world recognition~\cite{bendale2015towards,cen2022openworld} labels detected unknowns and adds them to the training set.
Pham et al.~\cite{pham2018bayesian} grouped regions perceptually to known and unknown instances, exploiting edges, boxes, and masks.
Recently, several works tackled open-set semantic segmentation, telling apart unknown areas from known classes~\cite{jung2021sml,hong2022goss,grcic2022densehybrid}.
%In the context of semantic segmentation, an open-set method tells apart unknown regions from the set of known classes, without distinguishing instances. Recently, several works tackled this task.
Among those not learning from OOD data, DML~\cite{cen2021mmsp} uses metric learning and
%to improve the uncertainty estimate from the maximum probability of the \textit{softmax} output. 
SML~\cite{jung2021sml} acts as post-processing, standardizing the max \textit{logits}, improving the class distributions.
%standardizing the max \textit{logits} for each class, thereby aligning the distributions and having each maximum better representing its class.
Instead, our work separates unseen, unknown objects into instances and segments known areas, addressing the proposed holistic segmentation.
%, thus finding and separating individual instances of unseen unknown objects, on top of performing panoptic segmentation, without using any OOD information.
% uncertainty

\textbf{Open-set panoptic segmentation}
The pioneering OSIS~\cite{wong2020uberlidaropenps} was the first in this direction. Applied to LiDAR data, it exploits 3D locations to cluster unlabeled points into instances.
Later, EOPSN~\cite{hwang2021exemplar} extended Panoptic FPN~\cite{kirillov2019panopticfpn} and grouped its proposals into clusters. At training time, EOPSN clusters similar unlabeled objects across multiple inputs. When surrounded by known segments, it labels an unlabeled object and uses it to learn to segment its instances. Instead, DDOSP~\cite{xu2022openpano_twostage} uses a known-unknown class discriminator and class-agnostic proposals.
However, as these approaches were intended to re-identify already-seen unlabeled objects, they all rely on seeing unknown data at training time~\cite{wong2020uberlidaropenps,hwang2021exemplar,xu2022openpano_twostage}. They all cluster into instances what falls in the predicted \textit{void} class, learned as a fallback (i.e., must be in the training samples) and assumed to contain all unknowns. Since datasets are limited~\cite{lehner20223d}, by requiring to learn from unknowns and the \textit{void} class, existing works are not designed to deal with any completely unseen object, as their pool of identifiable unknowns is also limited~\cite{hwang2021exemplar,xu2022openpano_twostage}. For these reasons, they solve only part of the problem. Instead, by using no OOD data at training time and preventing learning priors for unknowns, the proposed holistic segmentation differs from the way open-set panoptic segmentation has been tackled so far~\cite{wong2020uberlidaropenps,hwang2021exemplar,xu2022openpano_twostage}. As shown in Figure~\ref{fig:setting}, our setting allows to segment without constraints and separate even unseen unknown objects.
In contrast to all existing approaches, the proposed \pname\ neither relies on seeing unknowns at training time nor learns the \textit{void} class. \pname\ is the first method to solve this unconstrained, assumptions-free holistic segmentation setting.
%we argue that seeing OOD data at training time is a strong limitation and an oversimplification of the open-set problem at hand, and rather shares similarities with a closed-set scenario.
%Thanks to estimating the uncertainty and learning instance-aware embeddings, our \pname\ framework clusters uncertain regions into instances of unknown objects.

%\section{Task and Method} \label{sec:method}

%\textbf{Overview}
%In this work, we propose the task of holistic segmentation (Section~\ref{sec:holistic_task}) and address it with \pname\ (Section~\ref{sec:framework}).

\begin{figure*}[t]
\begin{center}
\includegraphics[width=1.00\textwidth]{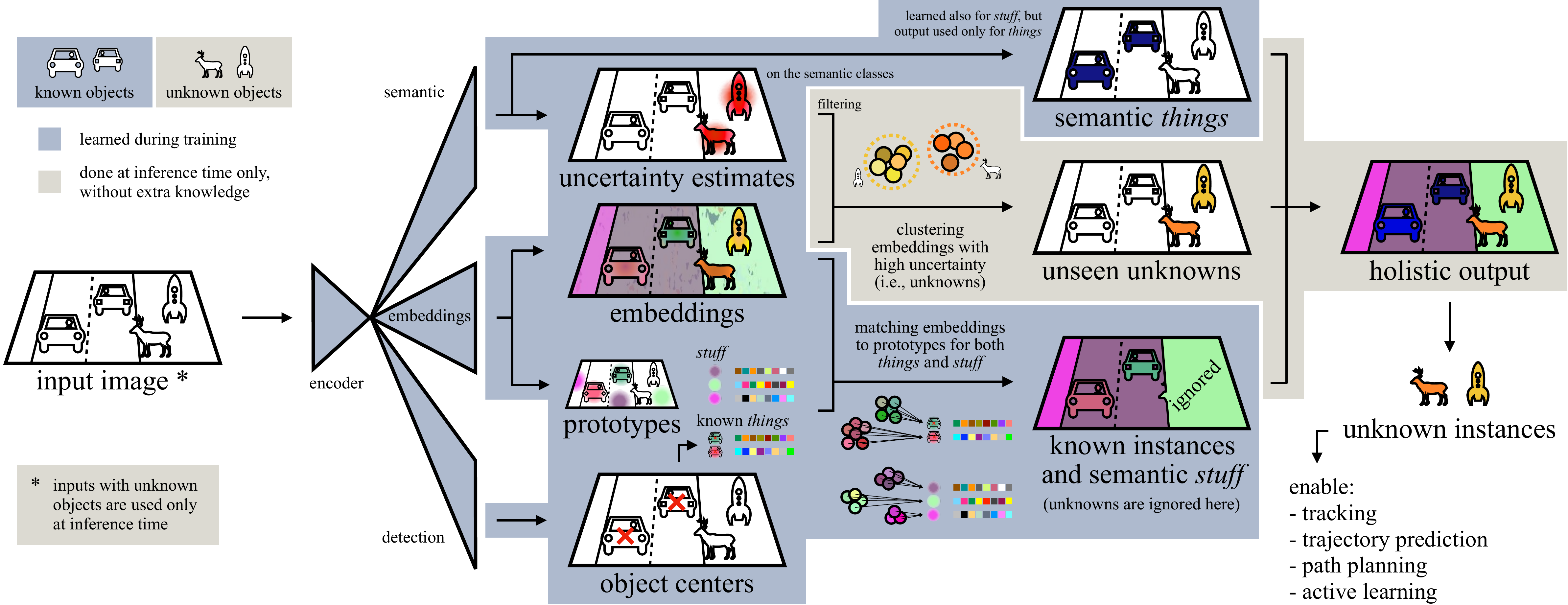}
\vspace{-0.7cm}
\end{center}
   \caption{The proposed \pname\ framework. Uncertainty is estimated in the semantic branch, and with the instance-aware embeddings, it determines unknown instances. Known instances are found via center regression and formed by grouping embeddings with their prototypes.}
\label{fig:framework}
\vspace{-0.2cm}
\end{figure*}

\section{Proposed Setting: Holistic Segmentation}
\label{sec:holistic_task}
As shown in Figure~\ref{fig:setting}, the proposed setting of holistic segmentation is a logical extension of open-set panoptic segmentation~\cite{wong2020uberlidaropenps,hwang2021exemplar}.
%The difference concerns unknowns and the way they are handled.
%We aim at broadening the problem to more realistic scenarios.
To make the setting unconstrained as real-life scenarios, we aim to identify and separate \textbf{any unseen, unknown object} into instances while segmenting known classes.
Other settings allow to include unknowns in the training data~\cite{wong2020uberlidaropenps,hwang2021exemplar,xu2022openpano_twostage} (e.g., within the \textit{void} class) or use information about them~\cite{zheng2021zero_instance}, and only re-identify already-seen unlabeled objects~\cite{hwang2021exemplar}.
Instead, we focus on the case where \textbf{no information is available about unknowns}.
%Instead of including unknowns in the training data (e.g., within the \textit{void} class) or using prior information about them, and only re-identifying already-seen unlabeled objects~\cite{hwang2021exemplar}, we want to identify any unseen unknown objects without learning from unknowns, nor using any information about them.
Therefore, holistic segmentation is more challenging, makes no assumptions about the training data (e.g., the presence of \textit{void}, with unknowns in it), leaves the problem unconstrained to any object, and simplifies data collection as no unknowns need to be in the training data. Formal definition and metrics follow open-set panoptic segmentation~\cite{wong2020uberlidaropenps,hwang2021exemplar}. As shown in Figure~\ref{fig:setting}, the outputs are comparable, but the definition of unknowns differs (here unseen), as well as the inability to learn from unknowns. Unknown instances can then be used for downstream tracking, trajectory prediction, path planning, or active learning.

\section{Proposed Framework: \pname}\label{sec:framework}
In Figure~\ref{fig:framework}, we show a representation of \pname, targeting holistic segmentation.
%(Section~\ref{sec:holistic_task}). 
\pname\ outputs instances of unseen unknowns by clustering instance-aware embeddings corresponding to highly uncertain regions (Section~\ref{sec:open_pano}). We use uncertainty estimation to distinguish known classes from unknowns, while embeddings are learned solely on known objects with panoptic segmentation (Section~\ref{sec:closed_pano}).
%We learn descriptive instance-aware embeddings solely on known objects (Section~\ref{sec:closed_pano}) and exploit uncertainty estimation to distinguish known regions from unknowns. Then, by clustering the learned embeddings associated to highly uncertain areas, we form instances of unseen unknowns (Section~\ref{sec:open_pano}).

\subsection{Panoptic Segmentation for Known Classes}\label{sec:closed_pano}
Our approach for closed-set panoptic segmentation builds upon learning instance-aware embeddings. As shown in Figure~\ref{fig:framework}, an encoder extracts features from an input image and propagates them to different decoders: 1) a semantic branch performing semantic segmentation and uncertainty estimation to identify unknown regions (Section \ref{sec:open_pano}); 2) a detection branch identifying object centers similarly to Panoptic-DeepLab~\cite{cheng2020panopticdeeplab}; and 3) an embeddings branch, with two separate heads, for prototypes and embeddings.

We make the embeddings instance-aware via discriminative loss functions (Section~\ref{sec:losses}) and by concatenating the detection branch features to prototype and embeddings heads. Embeddings and detections are made also semantic-aware by concatenating the semantic \textit{logits} to the last layers of the heads.
Prototypes $\Omega$ are feature vectors the prototype head predicts to represent objects and \textit{stuff} classes. While computed at each pixel, only the features at object centers $C$ are considered \textit{thing} prototypes, plus one for each \textit{stuff} class.
%The prototype head predicts a feature vector for every pixel. For \textit{things}, these are considered meaningful instance prototypes only at the corresponding object centers found by the detection branch. 
This is inspired by the size heatmap in~\cite{zhou2019objects}.
%Such \textit{thing} prototypes represent each instance, distinguishing them from one another. 
\textit{Thing} prototypes are $\Omega_{th} = \{(\mu_o, \sigma_o^2) \in \mathbb{R}^F \times \mathbb{R}^+ : o \in I\}$, one for each object $o$ of all detected instances $I$. $F$ is the embedding size, $\mu$ and $\sigma^2$ are mean and variance.
\textit{Stuff} prototypes are $\Omega_{st} = \{(\mu_k, \sigma_k^2) \in \mathbb{R}^F \times \mathbb{R}^+: k \in K_{st}\}$, one for each \textit{stuff} class $k \in K_{st}$ independently.

%For $\Omega_{th}$ only the values at the detected centers are extracted.
Simlarly to~\cite{wong2020uberlidaropenps}, the embeddings head predicts embeddings $\phi_{(i,j)} \in \mathbb{R}^F$ for each pixel $(i,j)$, then matches them with their prototype $\omega \in \Omega$ as follows.
We compute association scores $\hat{\mathbf{y}}_{(i,j),\omega}$ for each pixel $(i,j)$ and prototype as:
\begin{equation}
    %\hat{y}_{(i,j),\omega} = - \tfrac{||\phi_{(i,j)} - \mu_\omega||^2}{2*\sigma_\omega^2}
    \hat{\mathbf{y}}_{(i,j),\omega} = - {||\phi_{(i,j)} - \mu_\omega||^2}/{2\sigma_\omega^2}
    \label{eq:embed_assoc}
\end{equation}
Compared to~\cite{wong2020uberlidaropenps}, we relax the problem by not including the term $-\tfrac{F}{2} \log \sigma_\omega^2$, and let the embedding variance be indirectly controlled by the final task, which naturally bounds it (shown empirically in Section~\ref{sec:quant_results}).
% This simplification takes into account the high diversity of image pixels and features, compared to the more numerical point cloud geometries for which the term was proposed~\cite{wong2020uberlidaropenps}.
Then, we keep the prototype variance $\sigma_\omega^2$ strictly positive by using \textit{softplus}.

At inference time, for \textit{things}, the semantic class of each instance is determined by majority voting of its semantic branch predictions, ensuring output consistency. Instead, the ID is computed from the highest score in Eq.~\ref{eq:embed_assoc}.
For \textit{stuff} regions, we follow~\cite{wong2020uberlidaropenps}, determining the semantic classes by associating the pixel embeddings to the prototypes $\Omega_{st}$ via the highest scoring class from Eq.~\ref{eq:embed_assoc}.
This decoupling allows semantic awareness throughout the model.

\subsection{Dealing with Unseen Unknown Objects}\label{sec:open_pano}
We find unknown segments by relying on uncertainty estimates, which can help identify the knowledge boundaries of a model~\cite{sensoy2018dpn,liu2020sngp}. Specifically, instead of predicting the \textit{void} class and searching in it for unknowns as in~\cite{wong2020uberlidaropenps,hwang2021exemplar}, we estimate the uncertainty related to the semantic segmentation predictions and consider as unknown the areas with a high associated uncertainty. Although our framework can flexibly work with various uncertainty estimators (Section~\ref{sec:quant_results}), here we exemplify it with DPN~\cite{sensoy2018dpn,joo2020beingbayesian}, which we extended from image classification to semantic segmentation, and also improved its convergence in this context. We chose DPNs as they allow for minimal modifications at training time, i.e., replacing the \textit{softmax} with a strictly positive activation function while providing good uncertainty estimates on OOD data without training on such data~\cite{sensoy2018dpn}.

Following \cite{sensoy2018dpn}, we consider the evidence $e_k=\alpha_k -1$ as a measure of the number of hints given by data for a pixel to be assigned to a class $k \in K$ known classes, with $\alpha_k$ being the parameters of the Dirichlet distribution $Dir(\alpha)$.
We compute the uncertainty as $u = K /\sum^K _{k=1} \alpha_k$.
Given that the class probabilities $\mathbf{p} = \left \{ p_k: k=[1,...,K] \right \}$ follow a simplex (i.e., are positive and sum to 1), the class assignment corresponds to a Dirichlet distribution parametrized over the evidence, as the probability density function:
\begin{equation}
    \begin{split}
    & D(\mathbf{p}|\bm{\alpha}) = \begin{array}{l} B(\bm{\alpha})^{-1} \prod_{k=1}^K p_k^{\alpha_k -1} \end{array}
    \\
    & \text{with:} ~~
    B(\alpha) = \begin{array}{l} \prod_{k=1}^K \Gamma(\alpha_k) / \Gamma \left( \sum_{k=1}^K \alpha_k \right) \end{array}
    \end{split}
\end{equation}
where $\Gamma$ is the gamma function and $B(\alpha)$ is the $K$-dimensional multinomial beta function~\cite{sensoy2018dpn}.

We apply this to semantic segmentation by predicting a concentration parameter $\alpha^{(i,j)}$ for each pixel $(i,j)$, replacing the last layer with the smooth \textit{softplus} activation function, thus converting the \textit{logits} to a strictly positive vector, which we use as evidence $e^{(i,j)}$ in the Dirichlet distribution.
We learn this distribution with the semantic loss $\mathcal{L}_s$ minimizing the negative expected log likelihood of the correct class $Y^{(i,j)}$, for the random variable $\mathcal{X}^{(i,j)} \sim \text{Dir}(\alpha^{(i,j)})$:
\begin{equation}
\begin{split}
    \mathcal{L}_s^{(i,j)} &= -E[\text{ln}\ \mathcal{X}^{(i,j)}_{Y^{(i,j)}}]
    \\ &= \psi \left( \begin{array}{l}\sum_{k=1}^K \alpha_{(i,j),k} \end{array} \right) - \psi(\alpha_{Y^{(i,j)}})
\end{split}
    \label{eq:dpn_loss}
\end{equation}
where $\psi$ is the digamma function (i.e., $\Gamma$'s logarithmic derivative) and $\alpha_{(i,j),k}$ is the output of the semantic branch. Due to the difficulty of modeling the target distribution in our holistic setting, we omit the KL term used in~\cite{sensoy2018dpn}, simplifying the loss design (Section~\ref{sec:quant_results}).
After training on the closed-set data, we consider all pixels $(i,j)$ with an estimated uncertainty $u_{(i,j)} \ge \mu + t\cdot\sigma$ as unknown regions with $\mu$ and $\sigma^2$ being mean and variance of the uncertainties of all training pixels, and $t$ being a hyperparameter.

\textbf{Separating unknowns}
After finding the unknown segments, we cluster their instance-aware embeddings trained only on known objects into individual unknowns using DBSCAN~\cite{ester1996dbscan}. We find the DBSCAN hyperparameters on the training closed-set data (Appendix). Finally, we re-assign the few DBSCAN's outliers to their originally predicted semantic class, thus ignoring their uncertainty estimates.
%in these few cases.

\subsection{Learning to Find Knowns and Unknowns}\label{sec:losses}
We train our models with a combination of four losses. The semantic branch is optimized with $\mathcal{L}_s^{(i,j)}$ (Eq.~\ref{eq:dpn_loss}) over the whole image sized $W \times H$ as:
\begin{equation}
\begin{split}
    \mathcal{L}_{s} &= \begin{array}{l} \tfrac{1}{WH} \sum_{i,j} -E[\text{ln}\ \mathcal{X}^{(i,j)}_{Y^{(i,j)}}] \end{array} 
    \\ &= \begin{array}{l} \tfrac{1}{WH} \sum_{i,j} \psi \left( \sum_{k=1}^K \alpha_{(i,j),k} \right) - \psi(\alpha_{Y^{(i,j)}}) \end{array}
\end{split}
\end{equation}
As in~\cite{cheng2020panopticdeeplab}, the detection branch is trained with an L2 loss between predicted $\hat{C}$ and ground truth $C$ center heatmaps:
\begin{equation}
    \mathcal{L}_{o} = \begin{array}{l} \tfrac{1}{WH} \sum_{i,j} \left (\hat{C}^{(i,j)} - C^{(i,j)} \right) ^2 \end{array}
\end{equation}

For \textit{stuff}, we use the predicted $\Omega_{st}$ as a pseudo label to learn the prototypes $\Omega$. For \textit{things}, the same is done with $\Omega_{th}$ at the true instance centers. The prototype loss $\mathcal{L}_p$ is the cross-entropy on the \textit{softmax} of the association scores $\hat{\mathbf{y}}_{(i,j),\omega}$, as $\hat{\mathbf{z}}_{(i,j),\omega} = {\text{exp}(\hat{\mathbf{y}}_{(i,j),\omega})} / {\sum_{\omega' \in \Omega} \text{exp}(\hat{\mathbf{y}}_{(i,j),\omega'})}$, with $\omega_{(i,j)}$ being the pseudo label prototype:
\begin{equation}
    \mathcal{L}_{p} = \begin{array}{l} \tfrac{1}{WH} \sum_{i,j} -\text{log}(\hat{\mathbf{z}}_{(i,j),\omega_{(i,j)}}) \end{array}
\end{equation}

We learn embeddings $\phi_{(x,y)}$ with a discriminative loss~\cite{de2017discriminative} $\mathcal{L}_d$ (Appendix).
%, $\mathcal{L}_d$ is composed of three terms: variance $\mathcal{L}_{va}$, distance $\mathcal{L}_{di}$, and regularization $\mathcal{L}_{re}$.
%\begin{equation}
%    \mathcal{L}_{d} = \lambda_{41} \mathcal{L}_{va} + \lambda_{42} \mathcal{L}_{di} + \lambda_{43} \mathcal{L}_{re}
%\end{equation}
%weighted $\lambda_{41}=\lambda_{42}=1$ and $\lambda_{43}=0.001$.
The overall training objective is:
\begin{equation}
    \mathcal{L} = \lambda_{1} \mathcal{L}_{s} + \lambda_{2} \mathcal{L}_{o} + \lambda_{3} \mathcal{L}_{p} + \lambda_{4} \mathcal{L}_{d}
\end{equation}
%is the training objective, which is weighted as $\lambda_{1}=\lambda_{3}=\lambda_{4}=1$ and $\lambda_{2}=200$.

\section{Experiments and Results}
\label{sec:exp_results}

\subsection{Experimental Setup} \label{sec:setup}

\textbf{Datasets}
We conducted our experiments on three public datasets, namely Cityscapes~\cite{cordts2016cityscapes}, Lost\&Found~\cite{pinggera2016lost}, and MS COCO~\cite{lin2014coco}.
\textbf{Cityscapes} is a popular outdoor benchmark. Recorded around 50 different cities, mainly in Germany, it contains 19 classes: 8 \textit{things} and 11 \textit{stuff}. We followed the standard split, with 2975 images for training and 500 as validation set, reporting all metrics on the latter.
Also recorded in Germany, the \textbf{Lost\&Found} dataset contains a variety of unusual OOD objects placed in the middle of the road. We selected it because: 1) it was recorded with the same sensor setup as Cityscapes, allowing seamless transfers and removing the need for fine-tuning; 2) it contains only real images; and 3) unlike similar datasets~\cite{blum2021fishyscapes,chan2021segmentmeifyoucan}, it provides instance annotations for unknowns. Therefore, it is a challenging complement to Cityscapes for holistic segmentation. We did not train on Lost\&Found, but used it only to evaluate models trained on Cityscapes. We report all metrics on the \textit{unknown} class of its 1202 test samples.
\textbf{MS COCO} is a challenging large-scale benchmark for general image understanding, as it includes a variety of scenarios from indoor to outdoor. The 2017 panoptic split contains
%118k images for training and 5k for validation, across 
80 \textit{thing} categories, and 53 \textit{stuff} classes. We followed EOPSN~\cite{hwang2021exemplar} by treating as unknown the least frequent 20\% \textit{thing} classes (e.g., \textit{bear}, \textit{frisbee}). However, instead of turning their segments into \textit{void} and keeping their images in the training set as in~\cite{hwang2021exemplar}, we removed their samples completely and regarded them as unseen unknowns. This reduced the training samples to 98112, with 117 classes to learn. We report on the 827 validation samples with unseen classes.

\textbf{Evaluation metrics}
We evaluated the panoptic quality (\textbf{PQ}) metric~\cite{kirillov2019panoptic} separately for known classes and unknowns, including recognition (RQ) and segmentation (SQ) qualities.
%RQ is the popular F$_1$ score used in object detection, and SQ is the IoU of matched segments.
We report PQ on the held-out classes of COCO~\cite{lin2014coco}, the unknown class of Lost\&Found~\cite{pinggera2016lost}, as well as on the 19 known classes of Cityscapes~\cite{cordts2016cityscapes} for both \textbf{open and closed} settings. Specifically, in open cases, models detect both knowns and unknowns, while in closed settings, the same models predict only knowns, which in practice means ignoring the uncertainty estimates. By analyzing both, we explore the trade-off between detecting unknowns (open) and the in-domain performance (closed).

\textbf{Network architecture}
All our models share the structure with Panoptic-DeepLab~\cite{cheng2020panopticdeeplab}, using a ResNet50~\cite{he2016resnet} backbone and decoders following Deep-LabV3+~\cite{chen2018deeplab}.
ResNet50 was chosen to increase reproducibility with limited resources. As described in Section~\ref{sec:open_pano}, the only modification to the semantic decoder is applying the \textit{softplus} activation to quantify the uncertainty. The other branches follow Panoptic-DeepLab for detecting centers and DeepLabV3+ for the embeddings, with two heads.

\textbf{Implementation details}
For~\cite{cordts2016cityscapes,pinggera2016lost}, we used input images sized 1024$\times$512 and batch size 16. For~\cite{lin2014coco}, we fed 8 images sized 640$\times$480. We used the Adam optimizer until convergence, with an initial learning rate of 0.001, which was reduced by 2\% at each epoch. We set $t=3$ for the uncertainty threshold (i.e., 3 times the standard deviation) and $F=8$ for the embedding size to keep the memory low. We adjusted to the different data distribution of COCO with $t=1$. The backbone was pre-trained on ImageNet~\cite{deng2009imagenet}. The losses were weighted $\lambda_{1}=\lambda_{3}=\lambda_{4}=1$ and $\lambda_{2}=200$~\cite{cheng2020panopticdeeplab}.

\begin{table}[t]
\begin{center}
\begin{tabular}{ll|ccc}
\toprule

& & \multicolumn{3}{c}{Lost\&Found (\textit{unseen})} \\
Method & Assumptions & PQ & RQ & SQ \\

\midrule

EOPSN~\cite{hwang2021exemplar} & data, \textit{void} & 0$^*$ & 0$^*$ & 0$^*$ \\
OSIS~\cite{wong2020uberlidaropenps} & data, \textit{void} & 1.45 & 2.23 & \textbf{65.11} \\
U3HS~[ours] & \textbf{none} & \textbf{7.94} & \textbf{12.37} & 64.24 \\

\bottomrule
\end{tabular}
\end{center}
\vspace{-0.2cm}
\caption{Segmentation of unseen unknown objects (\textit{unknown} class) of \textbf{Lost\&Found}~\cite{pinggera2016lost} test set after training on Cityscapes~\cite{cordts2016cityscapes} and transferring with no fine-tuning. $^*$: EOPSN diverged (null TP).}
\label{table:lost_and_found}
\vspace{-0.2cm}
\end{table}

\textbf{Prior works}
%For a fair comparison, all methods were retrained with the setup described above.
We compared our \pname\ with open-set panoptic works: OSIS~\cite{wong2020uberlidaropenps}, which we adapted from LiDAR point clouds to images, EOPSN and its baselines~\cite{hwang2021exemplar}, and DDOSP~\cite{xu2022openpano_twostage}. Instead of training them directly on the unknown categories being evaluated (as in~\cite{hwang2021exemplar}), we followed their setup~\cite{wong2020uberlidaropenps,hwang2021exemplar,xu2022openpano_twostage} by training them with the \textit{void} class as fallback, and applied them to unseen unknowns. All methods followed this setup, except that ours ignored \textit{void}.
On COCO, we facilitated other works following the K=5\% setting of EOPSN~\cite{hwang2021exemplar}, thus turning 4 classes into \textit{void}, so they learned 4 classes less than ours.
We repurposed and extended a variety of uncertainty estimators~\cite{sensoy2018dpn,van2020uncertainty,liu2020sngp} from image classification to semantic segmentation (Appendix). We then extended them to holistic segmentation by incorporating them in our \pname\ framework.
%We refer to the Supplementary Material for additional details.

%\input{tables/pano_ablation}%--------------------
%\input{tables/pano_fused.tex}%--------------------

\subsection{Quantitative Results} \label{sec:quant_results}

\begin{table}[t]
\begin{center}
\begin{tabular}{ll|ccc}
\toprule

& & \multicolumn{3}{c}{COCO (\textit{unseen})} \\
Method & Assumptions & PQ & RQ & SQ \\

\midrule

EOPSN~\cite{hwang2021exemplar} & data, \textit{void} & 0.40 & 0.50 & \underline{80.30} \\
\textit{void}-train~\cite{hwang2021exemplar} & data, \textit{void} & 4.40 & 5.90 & 74.80 \\
\textit{void}-supp.~\cite{hwang2021exemplar} & data, \textit{void} & 4.50 & 6.00 & 75.90 \\
DDOSP~\cite{xu2022openpano_twostage} & data, \textit{void} & \underline{9.30} & \underline{11.20} & \textbf{82.50} \\
U3HS~[ours] & \textbf{none} & \textbf{9.62} & \textbf{13.20} & 72.84 \\

\bottomrule
\end{tabular}
\end{center}
\vspace{-0.2cm}
\caption{Segmentation of unseen, unknown objects (20\% least frequent, held-out classes) on the validation set of \textbf{MS COCO}~\cite{lin2014coco}. All others learn \textit{void} and are taken from~\cite{xu2022openpano_twostage}, added trailing zero.}
\label{table:coco}
\vspace{-0.2cm}
\end{table}
\begin{table*}[t]
\begin{center}
\begin{tabular}{l|l|ccc|ccc|ccc}
\toprule
& & \multicolumn{3}{c|}{Lost\&Found (\textit{unseen})} & \multicolumn{3}{c|}{open Cityscapes} & \multicolumn{3}{c}{closed Cityscapes} \\

ID & Method & PQ & RQ & SQ & PQ & RQ & SQ & PQ & RQ & SQ \\

\midrule

\multicolumn{1}{c|}{-} & Panoptic-DeepLab~\cite{cheng2020panopticdeeplab} & - & - & - & - & - & - & \underline{45.82} & \underline{57.66} & \underline{79.46} \\
\multicolumn{1}{c|}{-} & OSIS~\cite{wong2020uberlidaropenps} & 1.45 & 2.23 & 65.11 & 39.42 & 50.20 & 78.53 & 39.42 & 50.20 & 78.53 \\

\midrule

A1 & [ours] baseline: semantic uncertainty & 0.49 & 0.82 & 60.16 & 35.02 & 44.83 & 78.10 & 35.97 & 46.12 & 78.00 \\

A2 & A1 + relaxed embedding association & 3.64 & 5.27 & \textbf{69.09} & \textbf{42.14} & \textbf{53.46} & 78.83 & 43.99 & 55.98 & 78.59\\

A3 & A2 + prototype head = \textbf{\pname} & \textbf{7.94} & \textbf{12.37} & 64.24 & \underline{41.21} & \underline{51.67} & \underline{79.77} & \textbf{46.53} & \textbf{58.99} & 78.87\\

A4 & A3 -- reassigning outliers & \underline{7.85} & \underline{12.25} & 64.11 & 39.84 & 49.97 & 79.75 & \textbf{46.53} & \textbf{58.99} & 78.87 \\

A5 & A4 -- majority voting & \underline{7.85} & \underline{12.25} & 64.11 & 23.94 & 30.15 & 79.41 & 26.77 & 33.86 & 79.06 \\

A6 & A4 -- semantic embeddings & 2.33 & 3.48 & \underline{67.01} & 35.16 & 43.34 & \textbf{81.13} & 35.92 & 44.30 & \textbf{81.07} \\

\midrule

%\begin{tabular}{ll|ccc|ccc|ccc}
%\toprule
%\multicolumn{2}{l|}{Method} & \multicolumn{3}{c|}{Lost\&Found} & \multicolumn{3}{c|}{open Cityscapes} & \multicolumn{3}{c}{closed Cityscapes} \\

%Framework & Uncertainty & PQ & RQ & SQ & PQ & RQ & SQ & PQ & RQ & SQ \\

%OSIS~\cite{wong2020uberlidaropenps} & - & 1.45 & 2.23 & \textbf{65.11} & 39.42 & 50.20 & 78.53 & 39.42 & 50.20 & 78.53 \\

U1 & \pname~[ours] + \textit{softmax} uncertainty & 0.10 & 0.20 & 51.45 & 39.70 & 50.77 & 78.20 & 45.12 & 56.83 & \textbf{79.40} \\
%EBOPS~\cite{wong2020uberlidaropenps} & / & / & / & / & / & / & / & / & / \\
U2 & \pname~[ours] + DUQ~\cite{van2020uncertainty} & 0.56 & 0.89 & 62.56 & \textbf{41.68} & \textbf{53.14} & 78.42 & 45.90 & 58.17 & 78.90 \\
U3 & \pname~[ours] + DPN~\cite{sensoy2018dpn} & 2.09 & 3.30 & 63.43 & 38.90 & 49.56 & 78.49 & 44.91 & 56.95 & 78.85 \\
U4 & \pname~[ours] + SNGP~\cite{liu2020sngp} & \underline{4.65} & \underline{7.57} & 61.49 & 41.02 & \underline{51.98} & \underline{78.91} & \underline{46.23} & \underline{58.56} & \underline{78.95} \\

A3 & \pname~[ours] + improved DPN~[ours] & \textbf{7.94} & \textbf{12.37} & \underline{64.24} & \underline{41.21} & 51.67 & \textbf{79.77} & \textbf{46.53} & \textbf{58.99} & 78.87 \\

\bottomrule

\end{tabular}
\vspace{-0.2cm}
\end{center}
\caption{Segmentation comparison of models trained on Cityscapes~\cite{cordts2016cityscapes} and transferred to the test set of Lost\&Found~\cite{pinggera2016lost} without fine-tuning. All were trained with the same constraints (e.g., ResNet50~\cite{he2016resnet}, small batch, and image sizes). 
An ablation study (A1-A6) shows the impact of the main components of \pname, with A3 being our full approach. A3 is paired with various uncertainty estimators (U1-U4).}
\label{table:pano_fused}
\vspace{-0.2cm}
\end{table*}%--------------------

\textbf{Unseen unknowns, L\&F}
Table~\ref{table:lost_and_found} compares our \pname\ with prior approaches when segmenting instances of unseen unknowns from Lost\&Found~\cite{pinggera2016lost}.
%on the challenging transfer to Lost\&Found~\cite{pinggera2016lost} after training on Cityscapes~\cite{cordts2016cityscapes}.
OSIS~\cite{wong2020uberlidaropenps} was the first to address the more limited open-set panoptic segmentation setting, followed by EOPSN~\cite{hwang2021exemplar}. However, OSIS performance fell short on PQ for unseen unknowns, proving the severe limitation of relying on unknowns at training time. By learning \textit{void}, OSIS achieved the highest SQ, which ignores wrong predictions~\cite{kirillov2019panoptic}.
Instead, despite numerous attempts, EOPSN~\cite{hwang2021exemplar} did not work: it diverged as soon as the exemplars were mined, obtaining 0 true positives (TP). We attribute this to the inconsistent similarities within the \textit{void} class of Cityscapes, compared to those across existing major classes treated as \textit{void} (e.g., \textit{car} in their setup). This prevented EOPSN from forming meaningful clusters from the proposal features during training~\cite{hwang2021exemplar}. Despite the similar setup to EOPSN~\cite{hwang2021exemplar}, OSIS~\cite{wong2020uberlidaropenps} could converge since it does not rely on associating unknowns across images. Our \pname\ outperformed OSIS by 5.5 times on PQ.

\textbf{Unseen unknowns, COCO}
In Table~\ref{table:coco}, we show results on the 16 held-out categories of MS COCO~\cite{lin2014coco}. In this case, other works were trained following the K=5\% setup of EOPSN~\cite{hwang2021exemplar}, where 4 classes were learned as \textit{void} (\textit{car}, \textit{cow}, \textit{pizza}, and \textit{toilet}). While this allows them to learn more meaningful representations of unknowns (as unlabeled), it limits the number of classes they can distinguish semantically, e.g., they cannot identify cars. For the other works, the benefit of expanding the \textit{void} distribution by enforcing the inclusion of a variety of recurring objects (e.g., pizza and cars) is evident as it allowed EOPSN to converge, although to a low PQ on the unseen objects. DDOSP~\cite{xu2022openpano_twostage} delivered a PQ similar to ours, albeit requiring to turn some knowns into \textit{void}, learning unknowns via \textit{void} and only 113 classes. Without altering the data nor making any assumption on the training samples (e.g., the presence of unknowns within \textit{void} as in~\cite{hwang2021exemplar,xu2022openpano_twostage,wong2020uberlidaropenps}), our \pname\ performed the best on the unseen categories, especially on RQ (i.e., ability to form instances of unknowns), while learning the whole set of 117 classes, thereby distinguishing even more classes than the other works.
Avoiding data assumptions with respect to unknowns made our \pname\ effective at segmenting unseen unknowns across both datasets.

\textbf{Known-unknown}
%open and closed Cityscapes Table 1, Table 3
Table~\ref{table:pano_fused} reports the performances in-domain, under open and closed settings (Section~\ref{sec:setup}). Ideally, a method would suffer from no decrease in PQ between the two settings, meaning that its estimates are aligned with the distribution shift between knowns and unknowns. OSIS~\cite{wong2020uberlidaropenps} does not use uncertainty estimation, so it does not have these two operating modes, resulting in identical open and closed-set outputs, as if it had only the open setting (via the prediction of \textit{void}). Conversely, all others suffered from a reasonable decrease when extended to open-set. DUQ~\cite{van2020uncertainty} had the smallest gap, which could be attributed to its underestimation of the uncertainty, as supported by its low scores on Lost\&Found.
%(also in Table~\ref{table:sem_uncert}).

\textbf{Closed-set}
In Table~\ref{table:pano_fused}, we also compare our \pname\ with Panoptic-DeepLab~\cite{cheng2020panopticdeeplab}. For a fair comparison, both approaches and all others were trained with the same backbone, image, and batch sizes (Section~\ref{sec:setup}). As these were all smaller than those used in~\cite{cheng2020panopticdeeplab} due to the limited resources used, they resulted in a lower PQ than that reported in~\cite{cheng2020panopticdeeplab}. Nevertheless, our full approach (A3) achieved a slightly higher PQ on Cityscapes under the same setting. We attribute this to the effectiveness of the instance-aware discriminative embeddings learned by our approach, compared to the offset vectors and grouping used by Panoptic-DeepLab. As the focus is unknowns, experiments with improved training resources are out of the scope of this work.

\textbf{Uncertainty} In Table~\ref{table:pano_fused}, we compare various uncertainty estimations paired to our \pname\ framework (U1-U4, A3). While DUQ~\cite{van2020uncertainty} and \textit{softmax} underperformed compared to OSIS~\cite{wong2020uberlidaropenps}, DPN~\cite{sensoy2018dpn} and SNGP~\cite{liu2020sngp} achieved a higher PQ. Nevertheless, our improved DPN paired with our framework outperformed prior methods by a substantial margin (A3).
For DPN and SNGP, this can be attributed to the superiority of our uncertainty estimates. Compared to OSIS and EOPSN, \pname's combination of uncertainty estimation with instance-aware embeddings was more effective than learning \textit{void} when encountering wholly new and unseen objects, such as those found in unconstrained settings (e.g., this transfer to Lost\&Found).

\begin{figure*}[t]
\begin{center}
\includegraphics[width=1.00\textwidth]{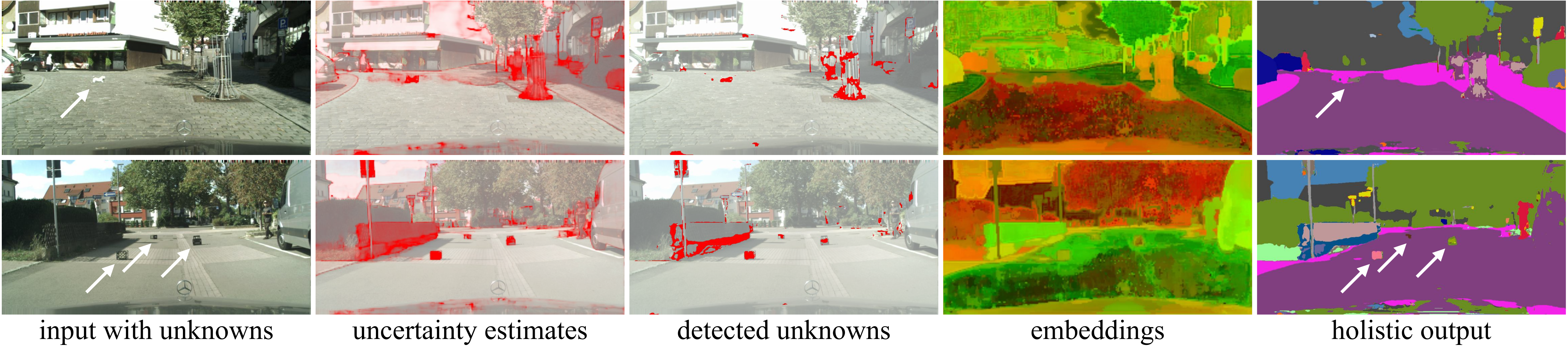}
\vspace{-0.7cm}
\end{center}
   \caption{Example predictions of \pname\ on OOD data from the Lost\&Found~\cite{pinggera2016lost} test set. The model was trained on Cityscapes~\cite{cordts2016cityscapes} and transferred to Lost\&Found without fine-tuning. Embeddings are projected to RGB via t-SNE~\cite{van2008tsne}. White arrows mark labeled unknowns.}
\label{fig:qualitative}
\vspace{-0.2cm}
\end{figure*}

\textbf{Ablation study}
%Table 3
Table~\ref{table:pano_fused} reports an ablation of the main components of our \pname, showing their benefits for holistic segmentation.
Compared to the open-set panoptic OSIS \cite{wong2020uberlidaropenps}, with A1, we reduced assumptions not learning the \textit{void} class, and we added a semantic branch with uncertainty for unknowns, which by itself worsened the performance.
However, combining this with a relaxed embedding association (Section~\ref{sec:closed_pano}) for \textit{things} and \textit{stuff} improved all metrics (A2). A dedicated prototype head (A3, i.e., full approach) increased them even further, more than doubling the PQ on unknowns (i.e., Lost\&Found). Specifically, dedicated heads allow both prototypes and embeddings to be more meaningful and expressive without sacrificing the other.
%below taken from the supplementary
A4 shows the impact of reassigning outliers (Section~\ref{sec:open_pano}). While its effect was limited on unknowns, it was more significant on Cityscapes~\cite{cordts2016cityscapes}. Transforming unknown predictions in standard in-domain outputs is relevant only in open settings. A5 shows the effect of majority voting to enforce consistency between the outputs (Section~\ref{sec:closed_pano}). This did not affect unknowns since classes are not distinguished among them, but it significantly impacted RQ and PQ on Cityscapes. Finally, A6 shows the importance of learning the embeddings according to their semantic classes. In A6, predictions are made by the dedicated semantic branch without the model learning to distinguish the embeddings semantically. Although this increased SQ, it caused a discrepancy within the model outputs, decreasing RQ and PQ.

\begin{figure}[b]
\vspace{-0.2cm}
\begin{center}
\includegraphics[width=1.00\linewidth]{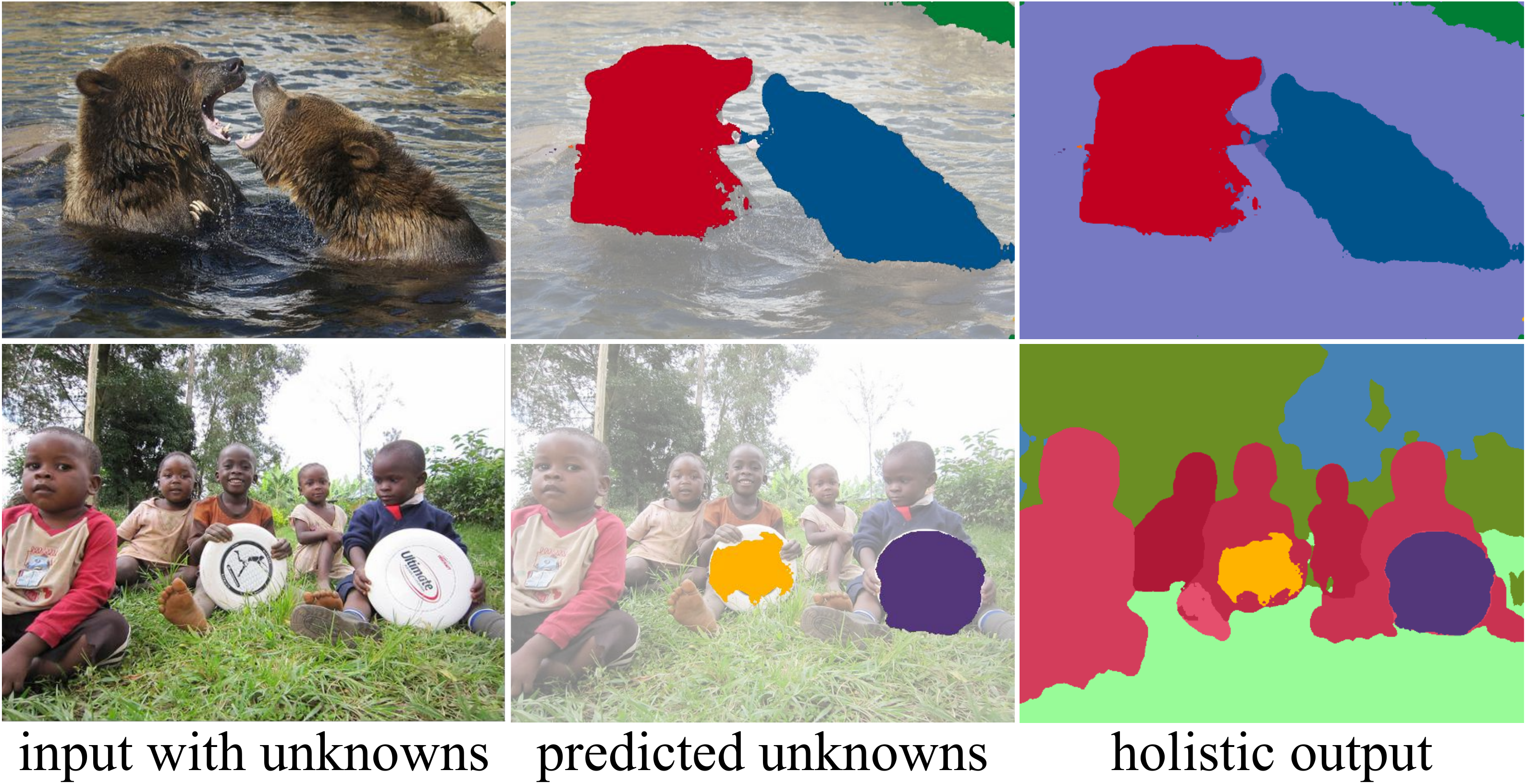}
\vspace{-0.7cm}
\end{center}
   \caption{Example predictions of \pname\ on OOD data from the COCO~\cite{lin2014coco} validation set. The model had never seen images containing \textit{bear} or \textit{frisbee} (part of the held-out classes), nor had any information about them. Colors represent the predicted instances.}
\label{fig:qualitative_coco}
%\vspace{-0.2cm}
\end{figure}

\subsection{Qualitative Results} \label{sec:qual_results}
Figure~\ref{fig:qualitative} shows example predictions of the proposed \pname. The images illustrate the setting difficulty and provide examples of the OOD objects of \textbf{Lost\&Found}~\cite{pinggera2016lost}. These are often small and hard to see, hidden in the shade or far away from the camera.
As seen in the quantitative results (Section~\ref{sec:quant_results}), our \pname\ could distinguish instances of unknowns (e.g., stroller in Figure~\ref{fig:teaser}), albeit leaving room for improvement. While unknowns correctly triggered high uncertainty estimates, their necessary filtering (third col.) sometimes left too few pixels, if any, on unknowns, leading to missed predictions. However, this is to be expected without any access to OOD data.
Furthermore, without distinguishing between unknown \textit{things} and unknown \textit{stuff}, also structures (e.g., fence in the lower image) were given an ID. Nevertheless, thanks to our learned instance-aware embeddings, these were not further subdivided but formed a single large instance (e.g., blue in the lower output). Separate unusual \textit{stuff} regions had the same effect, e.g., the structures around the trees in the upper image. This proves that instances are not simply created by separating disjoint OOD segments but are formed using the learned embeddings. As shown in Figure~\ref{fig:qualitative}, the embeddings are closely coupled with the uncertainty estimates and the outputs.

Figure~\ref{fig:qualitative_coco} reports predictions of \pname\ on samples of \textbf{MS COCO}~\cite{lin2014coco} containing two held-out classes (i.e., \textit{bear} and \textit{frisbee}). Remarkably, \pname\ was able to separate both bears and frisbees into individual instances despite their high inter-class similarity and not having accessed any information about them. This is thanks to the uncertainty estimation and instance-aware embeddings of our \pname.

\textbf{Data considerations and limitations}
Lost\&Found~\cite{pinggera2016lost} introduces a significant domain shift from Cityscapes.
By placing real OOD objects on the road, the authors had to choose unusual scenarios (Figure~\ref{fig:qualitative}), causing the whole scenes to be OOD. This leads to high uncertainty estimates also on a few known areas. As we do not use any OOD data, nothing constrains high uncertainty to unknown segments, decreasing PQ. A similar issue occurs in COCO, albeit less severely, thanks to more training data. 
However, COCO has no dedicated unknown class, so it had to be extracted from the set of known ones.
Nevertheless, results show that uncertainty is highly valuable, allowing to leave the settings unconstrained.
\pname~would mainly benefit from improvements in uncertainty estimates, embeddings descriptiveness, and their clustering. So, learning-based clustering~\cite{gasperini2021panoster,gasperini2020signal} could be advantageous.

The \textbf{Supplementary Material} includes more details on the proposed holistic segmentation setting, \pname\ and the baselines,
as well as additional results, including the trade-off between in-domain and OOD performances, failure cases and qualitative comparisons.

\section{Conclusion}
In this paper we introduced holistic segmentation: a new setting addressing completely unseen unknown objects in unconstrained scenarios. Additionally, we presented \pname:
the first solution for this new problem. Thanks to its uncertainty estimation and instance-aware learned embeddings, \pname\ identifies and separates instances of completely unseen unknowns without any information about them, while segmenting known regions.
Extensive experiments on multiple datasets showed the effectiveness of \pname.

%\clearpage

\begin{comment}
\bibliographystyle{ieee_fullname}
%\bibliographystyle{unsrt}
\bibliography{egbib}

\end{comment}

%\begin{comment}
%\clearpage
\appendix

\section{Supplementary Material}
In this appendix, we include further details and results. Specifically, Sections~\ref{sec:add_holistic}, \ref{sec:add_method} and~\ref{sec:add_implementation} provide deeper insights on the proposed setting, the method, and the experimental setup, respectively, while Sections~\ref{sec:add_quantitative} and~\ref{sec:add_qualitative} contain more results, both quantitative and qualitative.

\subsection{Additional Details on the Setting}\label{sec:add_holistic}
In this section, we describe the benefits of the proposed holistic segmentation setting in greater detail, considering both the impact on downstream tasks and the differences with other perception tasks addressing unknown objects.

The proposed holistic segmentation setting aims to segment any unseen, unknown objects without prior knowledge about the unknowns while segmenting known areas. In this context, "unseen unknowns" means any object of any category outside the known classes learned during training, such as the sheep in Figure~\ref{fig:sheep} for a method trained on, e.g., Cityscapes~\cite{cordts2016cityscapes}, as well as unidentified and distorted parts following a car accident.

\subsubsection{Motivation}
\begin{figure}[t]
\begin{center}
\includegraphics[width=1.00\linewidth]{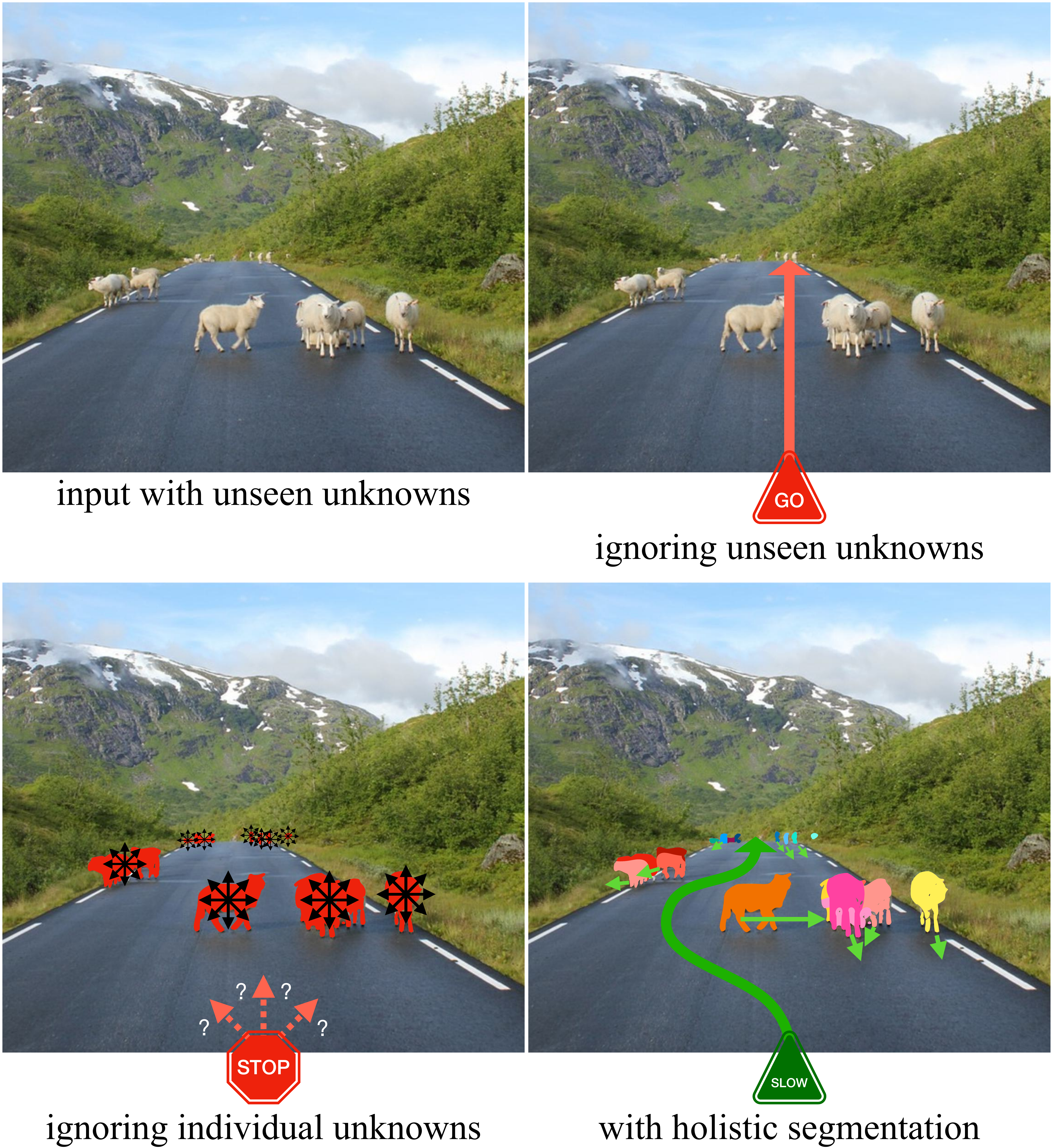}
\vspace{-0.7cm}
\end{center}
   \caption{Motivation diagram for identifying unseen, unknown objects on a sample of~\cite{chan2021segmentmeifyoucan} considering path planning as a downstream task, and hypothesizing that \textit{sheep} is not part of the training data (i.e., unseen unknown). Shown segments are not predictions.
   State-of-the-art approaches dangerously ignore unknowns (top right)~\cite{cheng2020panopticdeeplab}.
   %, which would lead the path planning module to output a dangerous straight path over the sheep.
   OOD segmentation does not identify instances of unknowns (bottom right)~\cite{jung2021sml}, making it difficult for downstream tasks as the unknowns cannot be tracked, and their trajectory cannot be predicted.
   %, preventing to understand their motion over time via tracking. This could lead to a stall as no path could be found.
   The proposed setting (bottom right) identifies individual unseen unknowns, %enabling tracking them and estimating their trajectories,
   leading to a safe path.}
\label{fig:sheep}
\vspace{-0.2cm}
\end{figure}

The importance of identifying unseen unknowns arises from safety-critical scenarios, such as autonomous driving, where ignoring them can lead to dangerous consequences when simply using the predicted segments for downstream tasks, e.g., path planning. This is shown in the top right of Figure~\ref{fig:sheep}.
Since even large-scale datasets are limited representations of the real world, there will always be corner cases and long tail samples which are problematic for standard models~\cite{lehner20223d}. Therefore, it is crucial to identify these cases and then deal with them safely via downstream tasks.

%\setfigurecounter{7}

\begin{table*}[t]
\renewcommand{\arraystretch}{1}
\begin{center}
\begin{tabular}{l|p{0.205\textwidth}p{0.205\textwidth}p{0.205\textwidth}}
\toprule

Setting & Data Assumptions & Identifiable Objects & Not Identifiable Objects \\

\midrule
\renewcommand{\arraystretch}{2}

open-set panoptic segm.~\cite{hwang2021exemplar,wong2020uberlidaropenps} & unknowns are already in the training data, within \textit{void} areas & known and unlabeled objects present in the training data as \textit{void} & categories outside of the training data~\cite{hwang2021exemplar} \\
&&&\\
open-vocabulary, zero-shot~\cite{xu2022open_voc,huynh2022openvocabulary} & the underlying language model knows about every unknown & objects known by the language model~\cite{radford2021clip} & categories outside of the training data of the language model~\cite{radford2021clip} \\
&&&\\
holistic segmentation [ours] & \textbf{none} & \textbf{any} known and unknown (i.e., \textbf{unseen}) object & \textbf{none} \\

\bottomrule
\end{tabular}
\end{center}
\vspace{-0.2cm}
\caption{Comparison of tasks and settings dealing with instances of unknown objects. The second column (Data Assumptions) is related to unknowns. The 2 rightmost columns represent the objects that are theoretically identifiable or not, given the setting.}
\label{table:setting_comp}
\vspace{-0.2cm}
\end{table*}
%\settablecounter{5}

OOD segmentation~\cite{jung2021sml}, i.e., segmenting unknown areas as a whole and not identifying individual instances of unknowns, flags the presence of something unknown in the input. In the context of downstream tasks, such as path planning, a single OOD segment (bottom left in the figure) could trigger an alert state, leading to a potential stop, which is a safe state. However, given that OOD segmentation does not separate unknown objects into instances, once found, it is unclear whether they are moving or static, which means that it would be difficult for path planning.
Instead, by segmenting instances of unseen unknowns (bottom right in the figure), holistic segmentation allows tracking unseen objects and estimating their trajectory, leading to a safe path. This motivates the instance segmentation of unknowns, which brings benefits similar to those of instance segmentation compared to semantic segmentation for known objects.

Also critical is the ability to deal with any unseen, unknown object category and not be restricted to a limited subset of them. This is of utmost importance to address the wide variability of objects and scenarios encountered in the real world. While previous settings focused on re-identifying already-seen objects~\cite{hwang2021exemplar,xu2022open_voc}, we design holistic segmentation specifically to address any unseen category. 

\subsubsection{Comparison with Other Settings}
As shown in Table~\ref{table:setting_comp}, compared with other tasks and settings also dealing with unknown objects, the proposed holistic segmentation makes no assumptions about the unknown objects, allowing one to segment any objects. Instead, zero-shot and open-vocabulary approaches assume that text descriptions of unknown objects are available~\cite{huynh2022openvocabulary,xu2022open_voc}. Open-set panoptic segmentation methods assume unknowns are confined within \textit{void} regions at training and test time~\cite{wong2020uberlidaropenps,hwang2021exemplar}. In the latter case, \textit{void} may not be available or not sufficiently large and diverse (as in Cityscapes~\cite{cordts2016cityscapes}), depending on the training data. Due to their construction, both of these setups inherently restrict the pool of recognizable objects to those for which text descriptions are available through a vision-language model (open-vocabulary) or to those present within their own training set (open-set panoptic).

For example, for the scene in Figure~\ref{fig:sheep}, because of its setup, EOPSN~\cite{hwang2021exemplar} cannot identify any \textit{sheep} unless a vast amount of images containing \textit{sheep} is part of its training data (with \textit{sheep} being labeled as \textit{void}, or directly as a dedicated class \textit{sheep}). Open-vocabulary methods would rely on the fact that a language model~\cite{radford2021clip} already knows about \textit{sheep} to be able to identify them in the image. While the concept of \textit{sheep} is relatively simple and could be assumed to be known by a large language model, there is no guarantee that such a model would know about every possible object and scene that can be encountered in real life (e.g., unidentified pieces on the road following a car accident), meaning that open-vocabulary approaches cannot deal with long tail samples from the distribution of the natural world, simply because their language model cannot process them.

Again, given that datasets include by definition only a fraction of the diversity of the world~\cite{lehner20223d}, also datasets to test the ability of a model to identify unknowns are limited~\cite{pinggera2016lost,chan2021segmentmeifyoucan,blum2021fishyscapes}, containing only a small amount of the possible objects and situations that can be encountered in real life. Therefore, to operate reliably in real unconstrained scenarios, it is of utmost importance not to have limitations on the types of recognizable objects, which should go beyond those found in existing datasets. Instead, relying on a language model to identify unknowns is equivalent to shifting the unknown problem to a different model. As shown with CLIP by Radford et al.~\cite{radford2021clip}, large language models also have issues with OOD samples. For example, unidentified broken car parts lying on the ground after an accident would be difficult to describe, so it would be problematic for language models. Thus, when given inputs that are unseen and unknown to the underlying language model, open-vocabulary and zero-shot methods would fail to identify the unknown objects. Furthermore, existing open-set panoptic works rely on the presence of unknowns (intended as unlabeled) directly in the training data through the \textit{void} class. This highlights the need for a new and unconstrained solution.

For these reasons, the critical differences between the proposed holistic segmentation setting and previous tasks are that holistic segmentation is not constrained in terms of the types of unknown objects that are identifiable and that holistic segmentation does not assume the presence of unknowns in the training data, thereby segmenting \textbf{any unseen, unknown object without any prior knowledge about unknowns}. Limited by design by either the unknowns that are known to the underlying language model (e.g., open-vocabulary) or the unknowns that are directly present in the training data (e.g., open-set panoptic segmentation), previous tasks do not enable the identification of any instance of unknowns and rely on prior knowledge about unknowns and their data distribution (e.g., through CLIP~\cite{radford2021clip} or by learning \textit{void}).

\subsection{Additional Details on the Method}\label{sec:add_method}

\textbf{Loss functions}
As described in Section~\reff{4.3}, the proposed method is trained with a combination of losses: a semantic loss $\mathcal{L}_s$, an object detection loss $\mathcal{L}_o$, a prototype loss $\mathcal{L}_p$, and a discriminative loss $\mathcal{L}_d$. %~\ref{sec:losses}
The discriminative loss is aimed at learning meaningful embeddings. It is composed of three different terms~\cite{de2017discriminative}, namely variance $\mathcal{L}_{va}$ to attract elements towards the mean, distance $\mathcal{L}_{di}$ to push away different groups, and regularization $\mathcal{L}_{re}$ to prevent the divergence of clusters from the origin:
\begin{equation}
    \begin{split}
    \mathcal{L}_{d} &= \begin{array}{l} \lambda_{41}\mathcal{L}_{va} + \lambda_{42}\mathcal{L}_{di} + \lambda_{43}\mathcal{L}_{re} \end{array} \\
    \mathcal{L}_{va} &= \begin{array}{l} \tfrac{1}{|\Omega|} \sum_{\omega \in \Omega} \tfrac{1}{N_\omega} \sum_{a=1}^{N_\omega} \left[ ||\mu_\omega -\phi_a|| - \delta_v \right]_+^2 \end{array}\\
    \mathcal{L}_{di} &= \begin{array}{l} \tfrac{1}{|\Omega|(|\Omega|-1)} \sum_{\omega_A \in \Omega} \sum_{\omega_B \in \Omega} \left[ 2 \delta_d - \right. \end{array}
        \\ & \qquad \qquad \qquad \qquad \qquad \quad  \begin{array}{l} \left. ||\mu_{\omega_A} - \mu_{\omega_B}||\right]_+^2 \end{array}\\
    \mathcal{L}_{re} &= \begin{array}{l} \tfrac{1}{|\Omega|} \sum_{\omega \in \Omega} ||\mu_\omega|| \end{array}
    \end{split}
\end{equation}
where: $|\Omega|$ is the number of prototypes, $N_\omega$ is the number of embeddings associated to the prototype $\omega$, $\mu_\omega$ is the mean embedding of the cluster related to $\omega$, $||\cdot||$ is the L2 distance, $\left[ x \right]_{+} = \textrm{max}(0,x)$ is the hinge (i.e., until which threshold the terms are active~\cite{de2017discriminative}), $\omega_A \neq \omega_B$, and we follow~\cite{de2017discriminative} for the hyperparameters, e.g., $\lambda_{41}=\lambda_{42}=1$ and $\lambda_{43}=0.001$.

\textbf{Clustering unseen unknowns}
As described in Section~\reff{4.2}, we use DBSCAN~\cite{ester1996dbscan} to cluster the embeddings of unknown regions into individual unknown objects. %~\ref{sec:open_pano}
Specifically, DBSCAN has multiple advantages: it does not need the number of clusters as input (which is unknown in our case), it is effective and very fast, has a low memory footprint, and distinguishes outliers (Table~\reff{3} shows the impact of this feature with A3-A4). %~\ref{table:pano_fused}
Although other traditional clustering methods (e.g., Mean Shift, Affinity Propagation, Birch) are theoretically applicable in our setting, they come with drawbacks (e.g., have high memory requirements, do not output outliers, are significantly slower, or tend to deliver sub-optimal results). On the other hand, popular approaches that require the number of clusters as input cannot be applied in our settings (e.g., K-Means). Hence, DBSCAN was selected.

\subsection{Additional Details on the Experimental Setup}\label{sec:add_implementation}

\textbf{Clustering parameters}
DBSCAN requires two parameters: $minPts$~(number of points in a neighborhood to count as a core point) and $\epsilon$ (maximum neighborhood size). To find such parameters, we trained a model (i.e., on Cityscapes~\cite{cordts2016cityscapes} or MS COCO~\cite{lin2014coco}), then selected ($minPts$, $\epsilon$) with a simple grid search maximizing PQ on a random subset of the known dataset (i.e., without unknowns). Towards this end, we formed instances as follows: ignoring the detection output (i.e., using only the embeddings) and determining their class via majority voting from the semantic output. In particular, when finding ($minPts$, $\epsilon$), this means treating the embeddings of knowns as if they were unknowns (apart from their semantic class), assuming that the model treats them similarly. It is essential to consider that the parameters were selected on the known objects (i.e., from Cityscapes or COCO), despite DBSCAN being used only for separating unknowns (i.e., in the Lost\&Found~\cite{pinggera2016lost} dataset or the held out classes of COCO). We did this to maintain the unknowns completely unseen (i.e., only as test set), as in real scenarios.

\begin{figure}[t]%{r}{7.5cm}
%\vspace{-0.55cm}
\begin{center}
\includegraphics[width=1.00\linewidth]{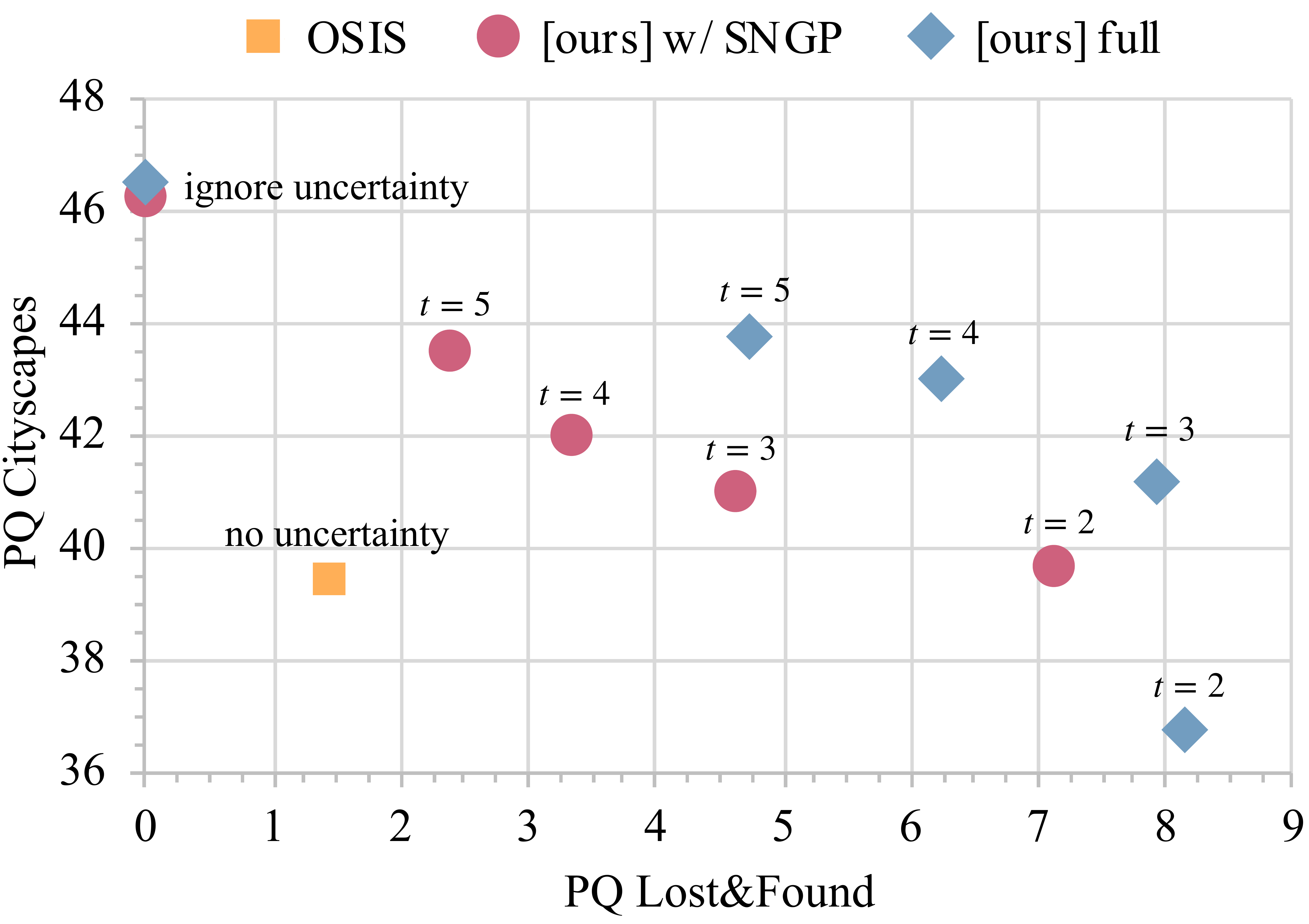}
\vspace{-0.7cm}
\end{center}
\caption{Trade-off between known (i.e., Cityscapes~\cite{cordts2016cityscapes} validation set) and unknown (i.e., Lost\&Found~\cite{pinggera2016lost} test set) performance introduced by OSIS~\cite{wong2020uberlidaropenps}, compared to our approach, both using SNGP~\cite{liu2020sngp} and our improved DPN (i.e., [ours] full, in blue). The different data points are obtained by varying the parameter $t$.
%the threshold $t$, as shown in the plot.%%%%%%%was this
}
\label{fig:plot}
%\vspace{-0.3cm}was this
\vspace{-0.2cm}
\end{figure}
%\setfigurecounter{8}

\textbf{Comparisons with previous works}
As described in Section~\reff{5.1}, we compared our uncertainty-based solution with prior works learning the \textit{void} class and tested various uncertainty estimators within our proposed framework. %~\ref{sec:setup}
The following paragraphs further detail how these other methods were trained.

Following our setup of training on Cityscapes~\cite{cordts2016cityscapes} and transferring to Lost\& Found~\cite{pinggera2016lost} without any fine-tuning, prior works addressing open-set panoptic segmentation (i.e., OSIS~\cite{wong2020uberlidaropenps} and EOSPN~\cite{hwang2021exemplar}) were trained by learning the \textit{void} class of Cityscapes, unlike our \pname. This unlabeled class comprises all pixels that do not fulfill the requirements to be part of one of the standard 19 annotated classes. Some of these \textit{void} pixels are systematic, e.g., the back side of traffic signs and street lights (excluding poles). By exploiting the variability within \textit{void}, the models learn the extra class decision boundary as a fallback covering anything far from the other classes. To do so, OSIS learns a constant $U$ representing such boundary. In particular, we adapted OSIS from LiDAR point clouds to RGB images, applying it to each pixel instead of point and changing architecture accordingly. As for ours and all other models in this work, we used a ResNet50~\cite{he2016resnet} as backbone and decoders following the structure of DeepLabV3+~\cite{chen2018deeplab}. Moreover, to keep the GPU memory low, we used the same $F=8$ for the embedding size as in our \pname.
For the experiments on MS COCO~\cite{lin2014coco}, we followed the K=5\% setup of EOPSN~\cite{hwang2021exemplar}, turning 4 classes into \textit{void} to facilitate prior works learning on \textit{void}, by ensuring a diverse distribution of its pixels, as they now cover a diverse set of classes (e.g., \textit{pizza} and \textit{car}).

\begin{table}[t]%{r}{7cm}
%\vspace{-0.1cm}
%\vspace{-0.8cm}
\begin{center}
\begin{tabular}{l|cc|c}
\toprule
& \multicolumn{2}{c|}{L\&F (\textit{unseen})} & open CS \\

Uncertainty method & AP & FPR$_{95}$ $\downarrow$ & mIoU \\

\midrule

\textit{softmax} & 16.72 & 22.88 & \textbf{71.77}  \\
MC Dropout~\cite{gal2016mcdropout} & 11.22 & \underline{13.94} & 68.31 \\
DML~\cite{cen2021mmsp} & 3.14 & 83.04 & 69.86 \\
DUQ~\cite{van2020uncertainty} & 5.43 & 26.64 & 68.78 \\
DPN~\cite{sensoy2018dpn} & 5.43 & 19.79 & 66.99 \\
SML~\cite{jung2021sml} & 16.91 & 51.67 & \underline{70.69} \\
SNGP~\cite{liu2020sngp} & \underline{22.70} & \textbf{12.02} & 70.68 \\
improved DPN [ours] & \textbf{25.44} & 19.10 & 70.10 \\

\bottomrule
\end{tabular}
\vspace{-0.2cm}
\end{center}
\caption{Comparison of open-set semantic segmentation on Lost\&Found~\cite{pinggera2016lost} test set of uncertainty estimators based on DeepLabV3+~\cite{chen2018deeplab} and trained only on Cityscapes (CS)~\cite{cordts2016cityscapes}.}
\label{table:sem_uncert}
\vspace{-0.2cm}
\end{table}
%\settablecounter{6}

For the other uncertainty estimation approaches evaluated in this work, we used the authors descriptions and implementations, adapting~\cite{van2020uncertainty,liu2020sngp} from image classification to semantic and panoptic segmentation. For DML~\cite{cen2021mmsp}, we used the authors best hyperparameters, therefore a variance loss weight $\gamma_{VL}=0.01$, and weights $\beta=20$ and $\gamma=0.6$. For SML~\cite{jung2021sml}, we did not employ the boundary suppression, as it did not improve the results. This might be due to Lost\&Found~\cite{pinggera2016lost} being annotated only for the OOD objects and a coarse road segment. For DUQ~\cite{van2020uncertainty}, we used an embedding dimension of $m=8$, due to constrained training resources, same as our $F=8$. Then, we used length scale $\sigma^2=0.3$ and exponential smoothing factor $\gamma=0.999$. For SNGP~\cite{liu2020sngp}, we again used an embedding dimension $D=8$ (due to the limited training resources), no layer norm for the embeddings, an exponential smoothing factor $\gamma=0.99$ for updating $\Sigma$ and 50 samples for Monte Carlo averaging to estimate the uncertainty.

\textbf{MS COCO} As described in the main paper, given that there is no official set of unknown classes for MS COCO~\cite{lin2014coco}, we treat as unknown the least frequent 20\% known classes. These classes are: \textit{baseball bat}, \textit{bear}, \textit{fire hydrant}, \textit{frisbee}, \textit{hair drier}, \textit{hot dog}, \textit{keyboard}, \textit{microwave}, \textit{mouse}, \textit{parking meter}, \textit{refrigerator}, \textit{scissors}, \textit{snowboard}, \textit{stop sign}, \textit{toaster}, and \textit{toothbrush}. We held out all training samples where any of these 16 classes appeared, such that they were completely unseen to the models.

\textbf{Ablation study for holistic segmentation} With reference to Table~\reff{3}, A1 is our baseline, which was built upon OSIS~\cite{wong2020uberlidaropenps}. %~\ref{table:pano_fused}
As OSIS, A1 included learned instance-aware embeddings, but unlike OSIS, it featured a semantic decoder delivering semantic segmentation and uncertainty estimates based on the semantic output via our improved DPN. Moreover, as for all our models, A1 did not learn the \textit{void} class (unlike OSIS). 
A2 featured the relaxed score for the embedding association (described in Section~\reff{4.1}), which lets the variance be indirectly controlled by the final task (i.e., the loss $\mathcal{L}_p$, Section~\reff{4.3}). %~\ref{sec:closed_pano} ~\ref{sec:losses}
Unlike A1 and A2, which had a shared head between embeddings and prototypes (i.e., as in OSIS), A3 introduced a dedicated prototype head. In practice, this meant having more layers fully dedicated to the embeddings and the prototypes separately instead of sharing the computation until a later stage. Therefore, this allowed for more expressive and purposed features.
A4 did not reassign to the known classes the outliers obtained from clustering unknowns via DBSCAN. Therefore, these pixels were kept unknown and shared the same instance ID.
A5 did not perform majority voting (Section~\reff{4.1}). %~\ref{sec:closed_pano}
This meant directly assigning the semantic classes predicted by the semantic branch to all known instance pixels instead of enforcing coherence within an instance. This caused the instances to be fragmented according to how many semantic classes they contained, decreasing RQ.
Finally, A6 predicted the semantic classes for \textit{stuff} areas directly from the semantic prediction branch instead of matching the embeddings with \textit{stuff} prototypes as in A1-A5 (Section~\reff{4.1}). %~\ref{sec:closed_pano}

\begin{table}[t]%{r}{7.3cm}
%\vspace{-0.8cm}
%\vspace{-0.1cm}
\begin{center}
\begin{tabular}{lll|cc|c}
\toprule
\multicolumn{3}{l|}{Configuration} & \multicolumn{2}{c|}{L\&F (\textit{unseen})} & open CS \\

Ref. & Activ.F. & KL & AP & FPR$_{95}$ $\downarrow$ & mIoU \\

\midrule

\cite{sensoy2018dpn} & \textit{exp} & yes & 5.43 & 19.89 & 66.99\\

[ours] & \textit{softplus} & yes & 3.43 & 25.97 & 64.36 \\

[ours] & \textit{softplus} & no & \textbf{25.44} & \textbf{19.10} & \textbf{70.10} \\

\bottomrule
\end{tabular}
\vspace{-0.2cm}
\end{center}
\caption{Ablation study on uncertainty estimates for open-set semantic segmentation. Models trained only on Cityscapes~\cite{cordts2016cityscapes}.}
\label{table:uncert_ablation}
\vspace{-0.2cm}
%\vspace{-0.2cm}
\end{table}
%\settablecounter{7}

\subsection{Additional Quantitative Results}\label{sec:add_quantitative}

\textbf{Trade-off between known and unknown}
Figure~\ref{fig:plot} shows the trade-off between the performance on known and unknown for our framework, both with SNGP~\cite{liu2020sngp} and our improved DPN, compared to that of OSIS~\cite{wong2020uberlidaropenps}. The different data points were extracted by evaluating the outputs at different thresholds $t$, namely $[2, ..., 5]$, and ignoring the uncertainty estimates entirely (i.e., closed-set, reported where PQ Lost\&Found is 0). The hyperparameter $t$ directly affects how high the uncertainty estimates must be for their associated pixels to be considered unknown. This has an impact on the performance on open-set Cityscapes~\cite{cordts2016cityscapes} and Lost\&Found~\cite{pinggera2016lost}, since changing in output what is considered unknown alters what is regarded as in-domain (i.e., known) as well. OSIS~\cite{wong2020uberlidaropenps} does not have such a hyperparameter as it considers unknown everything predicted as \textit{void}. Overall, it can be seen that our proposed framework offers a better trade-off in both configurations (red and blue) than that of OSIS~\cite{wong2020uberlidaropenps} (yellow). Furthermore, using our full approach (i.e., our framework with our improved DPN) typically gave the best trade-off between known and unknown without compromising the metrics too much (blue).

\textbf{Unknowns in semantic segmentation}
% Table 2
In Table~\ref{table:sem_uncert}, we compare the ability of a wide variety of uncertainty estimators (i.e., \cite{sensoy2018dpn,van2020uncertainty,liu2020sngp,jung2021sml,cen2021mmsp}, and MC Dropout with 25 runs~\cite{gal2016mcdropout}) to find unknowns in a semantic setting on Lost\&Found~\cite{pinggera2016lost}, after training on Cityscapes~\cite{cordts2016cityscapes}. This meant retraining all methods under the same conditions while also extending DPN~\cite{sensoy2018dpn}, DUQ~\cite{van2020uncertainty}, and SNGP~\cite{liu2020sngp} to semantic segmentation.
Semantic models (Tables~\ref{table:sem_uncert} and~\ref{table:uncert_ablation}) used smaller crops sized 512$\times$256 compared to the other experiments.
For uncertainty estimation, we evaluated the ability to identify unknowns reporting the AP on the unknown class~\cite{pinggera2016lost}, as well as the false positive rate at the recall 95 (FPR$_{95}$). For semantic segmentation on Cityscapes, we computed the mIoU.
%Although \textit{softmax} delivered a competitive AP on the unknown objects (Table~\ref{table:sem_uncert}), it severely underperformed in a panoptic setting (Table~\ref{table:pano_all}). We believe this is due to the importance of learning instance-aware embeddings and the synergy required across the branches, which fell short.
As seen in Table~\reff{3}, DUQ~\cite{van2020uncertainty} and DPN~\cite{sensoy2018dpn} performed worse than SNGP~\cite{liu2020sngp}. %~\ref{table:pano_fused}
MC Dropout~\cite{gal2016mcdropout} underperformed \textit{softmax}, probably due to the contrasting opinions from 25 forward passes. Our method was the best at finding unknowns (AP) with high-quality uncertainty estimates (FPR$_{95}$). Table~\ref{table:sem_uncert} also reports the mIoU on Cityscapes (CS), showing that all methods introduce a trade-off between OOD and in-domain outputs, as overestimating the uncertainty decreases the in-domain mIoU. Balancing these two complementary aspects is not trivial, with our approach and SNGP managing it best.

\begin{table}%{r}{4.2cm}
%\vspace{-0.8cm}
\begin{center}
\begin{tabular}{ll|ccc}
\toprule
ResNet depth & $F$ & PQ & RQ & SQ \\

\midrule

18 & 2 & 33.0 & 42.3 & 77.9 \\
18 & 4 & 38.9 & 49.8 & 78.0 \\
18 & 8 & 41.3 & 52.7 & 78.3 \\
18 & 16 & \textbf{42.3} & \textbf{53.8} & \textbf{78.7} \\
18 & 32 & \underline{42.1} & \underline{53.6} & \underline{78.5} \\

\midrule

50 & 8 & \textbf{47.7} & \textbf{60.4} & \textbf{79.0} \\

\bottomrule
\end{tabular}
\vspace{-0.2cm}
\end{center}
\caption{Different embedding dimensions $F$ on closed-set panoptic segmentation on the validation set of Cityscapes~\cite{cordts2016cityscapes}. The first column indicates the depth of the ResNet~\cite{he2016resnet} backbone used (i.e., 18 for ResNet18).}
\label{table:embeddings}
\vspace{-0.2cm}
\end{table}
%\settablecounter{8}

\textbf{Ablation on uncertainty estimation}
%Table 4
%softplus is smooth version or ReLU, but differentiable everywhere, monotonic, and grows slower than exp
Table~\ref{table:uncert_ablation} compares the DPN~\cite{sensoy2018dpn} we adapted from image classification to semantic segmentation with our extension.
%Analogously to CertainNet~\cite{gasperini2021certainnet} over the adaptation of DUQ~\cite{van2020uncertainty} to object detection, 
Our improvements were oriented to simplify the training process and help convergence.
First we applied the \textit{softplus} activation function to the last semantic layer, instead of \textit{exp} as in DPN~\cite{sensoy2018dpn}. We chose \textit{softplus} because it grows slower than \textit{exp} and it is smooth, differentiable everywhere, and monotonic. This significantly improved the training stability at the cost of a reduced quality of the uncertainty estimates.
%helping to contain the high gradients typical of unstable trainings. Although this reduced the uncertainty estimates quality, it majorly stabilized training.
Finally, due to the complexity of modeling the target distribution in our setting, omitting the KL term used by DPN~\cite{sensoy2018dpn} further stabilized training and boosted the performance on all metrics.

% table clustering was here for standalone supplementary

\begin{figure*}[t]
\begin{center}
\includegraphics[width=1.00\textwidth]{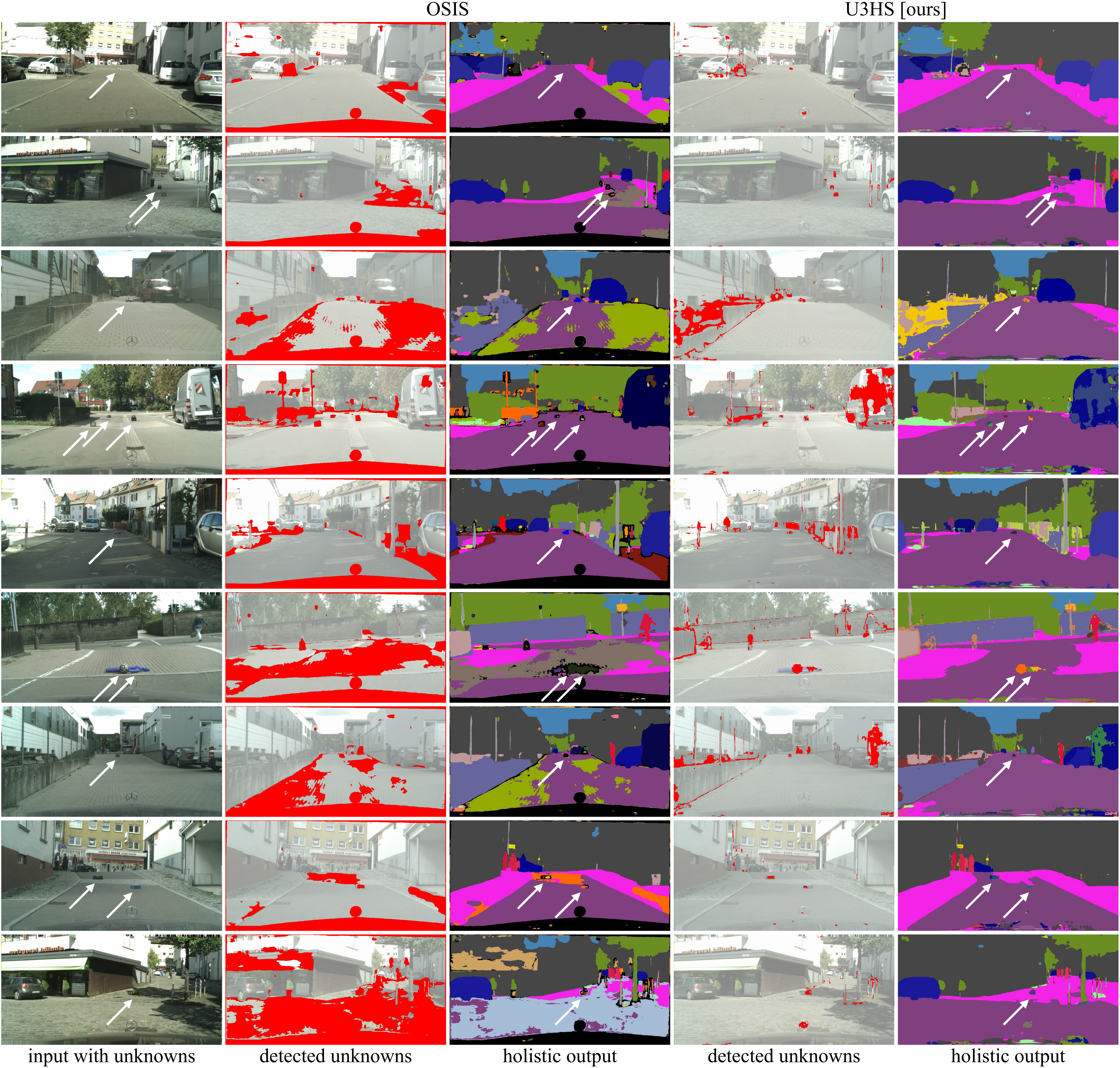}
\vspace{-0.7cm}
\end{center}
   \caption{Example predictions of OSIS~\cite{wong2020uberlidaropenps} and the proposed \pname\ on unknown categories from the test set of Lost\&Found~\cite{pinggera2016lost}. The models were trained on Cityscapes~\cite{cordts2016cityscapes} and transferred to Lost\&Found without any fine-tuning. OSIS found unknowns as the \textit{void} class (learned during training), while our \pname\ discovered them via uncertainty estimation. Black regions in OSIS's outputs, including around unknowns, represent pixels predicted as part of the unknown instance of the ego vehicle bonnet: since the bonnet is labeled as \textit{void} in the training set, OSIS learned it as such and it turned it into an unknown instance at inference time. White arrows mark labeled OOD objects.}
\label{fig:qualitative_comp}
\vspace{-0.2cm}
\end{figure*}
%\setfigurecounter{9}

\begin{figure*}[t]
\begin{center}
\includegraphics[width=1.00\textwidth]{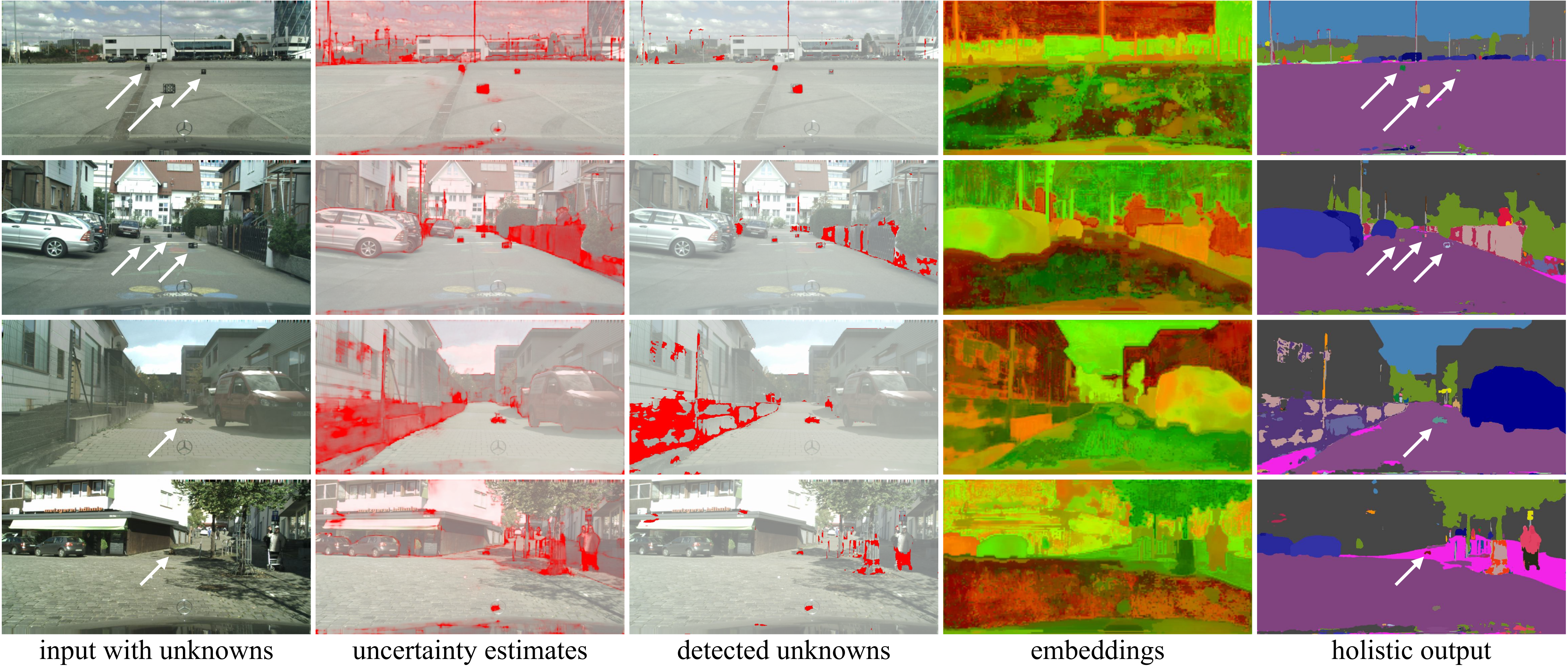}
\vspace{-0.7cm}
\end{center}
   \caption{Additional example predictions of the proposed \pname\ on unknown categories from the test set of Lost\&Found~\cite{pinggera2016lost}. The model was trained on Cityscapes~\cite{cordts2016cityscapes} and transferred to Lost\&Found without any fine-tuning. White arrows mark labeled OOD objects.}
\label{fig:qualitative_suppl}
\vspace{-0.2cm}
\end{figure*}
%\setfigurecounter{10}

\textbf{Impact of embedding size and architecture}
Table~\ref{table:embeddings} shows the effect of different embeddings dimensions $F$ on a smaller ResNet18~\cite{he2016resnet}. In the rest of this work, all experiments used $F=8$ and ResNet50, as in the last line of the table, due to constrained training resources. The embedding dimension directly affects the learning capability of the model. Since the instance-aware embeddings are a critical part of the output, a smaller $F$ is linked to inexpressive embeddings that cannot be as discriminative as those from a larger $F$.
Therefore, increasing $F$ improved all metrics except for the larger $F=32$. This can be attributed to the small ResNet18 backbone being already saturated at $F=16$, unable to extract rich and detailed features for the larger embeddings to exploit. With a larger model (e.g., ResNet101), even higher embedding dimensions $F$ might be beneficial. Table~\ref{table:embeddings} shows that our proposed approach, given less constrained resources, could deliver better results when using an embedding dimension higher than the $F=8$ employed across this work. The table also shows the comparison between ResNet18 and ResNet50, with the latter delivering over 15\% higher PQ at the same $F=8$. This shows how our proposed approach would perform with a larger backbone.

\begin{table}[t]
%\vspace{-0.1cm}
\begin{center}
\begin{tabular}{ll|ccc}
\toprule
%& & \multicolumn{3}{c}{Lost\&Found} \\

Method & Clustering & PQ & RQ & SQ \\

\midrule

\pname~[ours] & Mean Shift & 2.71 & 3.91 & \textbf{69.22} \\

\pname~[ours] & DBSCAN & \textbf{9.36} & \textbf{14.83} & 63.14\\

\bottomrule
\end{tabular}
\vspace{-0.2cm}
\end{center}
%\vspace{-1em}
\caption{Transfer from Cityscapes~\cite{cordts2016cityscapes} to Lost\&Found-300~\cite{pinggera2016lost} test set (i.e., on the first 300 samples, see Section~\ref{sec:add_quantitative}). DBSCAN (our choice) is compared to Mean Shift to cluster the embeddings of unknown areas.}
%\vspace{-0.65em}
\label{table:pano_cluster}
\vspace{-0.2cm}
\end{table}
%\settablecounter{9}

\textbf{Impact of the clustering method}
In Table~\ref{table:pano_cluster} we compare two popular clustering methods within our \pname\ framework, namely DBSCAN and Mean Shift~\cite{comaniciu2002meanshift}.
Due to the very high computation effort and memory required by Mean Shift, we opted for the following setup for this experiment. First, instead of a standard CPU implementation, we used a parallelized CUDA version of the algorithm~\cite{you2022gpu_meanshift}.
Then, due to the still very high memory requirements, specific samples of Lost\&Found caused memory issues. Therefore, we reduced the size of the test set of Lost\&Found~\cite{pinggera2016lost} to its first 300 samples (Lost\&Found-300), which were not problematic. These 300 samples are sufficient to indicate the effect of using Mean Shift instead of DBSCAN.
Table~\ref{table:pano_cluster} shows the superiority of DBSCAN for this setting, with a 3.5x higher PQ and 3.8x better RQ. In particular, RQ should be the focus as we compare instance segmentation of unknowns.

\textbf{Additional details on EOPSN vs.~OSIS} As described in Section~\reff{5.2}, EOPSN always diverged on Cityscapes despite numerous attempts, leading to null true positives, as shown in Table~\reff{1}. OSIS did not suffer from this issue: EOPSN's mining strategy requires associating similar "unknowns" across different inputs, but OSIS operates frame-by-frame. Since \textit{void} pixels are unstructured and undefined in Cityscapes, EOPSN's association fails. As this process is a fundamental step of its training procedure, it makes EOPSN diverge and leads to null scores due to the absence of true positives. Instead, in EOPSN's setup (i.e., re-identifying unlabeled objects seen during training), such associations can be made across the instances of the classes EOPSN's authors treat as \textit{void} or "unknown" (e.g., all cars in their setup, as in Table~\reff{2}). Therefore, on MS COCO (Table~\reff{2}), EOPSN managed to identify a few true positives, thereby scoring more than 0 for PQ and RQ and significantly more for SQ as it considers only the IoU of matched segments (TP) and not the wrong predictions (FP and FN).

\textbf{Varying number of unknowns}
Both EOPSN and DDOSP showed that their performance drops across the board by increasing the amount of \textit{void} classes to detect (i.e., $K$, pseudo-unknowns), especially for unknowns. As shown in Table~\reff{1}, they perform poorly also with $K=0$. Instead, our \pname\ does not rely on \textit{void}, so it is unaffected by $K$ or what is assigned to \textit{void}. This means that \pname's performance varies only slightly with different $K$s, as for random initialization.
Moreover, whether turning known classes into \textit{void} (MS COCO, Table~\reff{2}, with $K=5\%$) or not (Lost\&Found, Table~\reff{1}, $K=0$), our method outperforms prior works despite letting the others learn from unknowns via \textit{void}. Figure~\ref{fig:qualitative_comp} shows this qualitatively. Thanks to uncertainty estimation, our setup has an edge with unseen categories (e.g., Table~\reff{1}).
Increasing $K$ for open-set works (i.e., treating more classes as \textit{void}) means reducing the number of classes that can be segmented semantically.
Therefore, the proposed setting is more practical than open-set panoptic and open-vocabulary because, for ours, no unknowns need to be part of the training of any model (simpler data collection), and ours detects any unseen categories.

\subsection{Additional Qualitative Results}\label{sec:add_qualitative}

\subsubsection{Qualitative Comparison}
Figure~\ref{fig:qualitative_comp} shows a comparison of the predictions of the proposed \pname\ with the prior work OSIS~\cite{wong2020uberlidaropenps}, as well as the regions each predicted as unknown, on a set of samples from Lost\&Found~\cite{pinggera2016lost}. In particular, ours found unknowns as segments estimated as highly uncertain, and OSIS found them as the pixels predicted to be part of the learned \textit{void} class.

From the images, it can be seen how for the most part, OSIS managed to learn a relatively good class boundary around the \textit{void} class, as it was typically able to predict the OOD objects as unknown via \textit{void}. This is interesting as it shows how OSIS can potentially work with challenging unseen unknowns. However, the same figure also shows the substantial limitations of learning and predicting \textit{void} due to the assumptions about the data distributions this entails. In the first image, OSIS completely ignored the unknown object, assigning it to the \textit{road} class, while in the fifth image, it detected the toy as \textit{car}. In contrast, in the last picture, OSIS predicted almost everything as unknown. This proves how the binary aspect introduced by predicting the \textit{void} class (a pixel is either unknown, by being \textit{void}, or known, if another class) does not cope well with the diversity and unpredictability of the scenes in unconstrained real-world settings.
Specifically, predicting the \textit{void} class severely relies on the closed-set training data, as the success of such a method is directly related to the diversity of the \textit{void} class seen during training, which is limited as it cannot correctly sample the long tail of the data distribution~\cite{lehner20223d}.

\begin{figure*}[t]
\begin{center}
\includegraphics[width=1.00\textwidth]{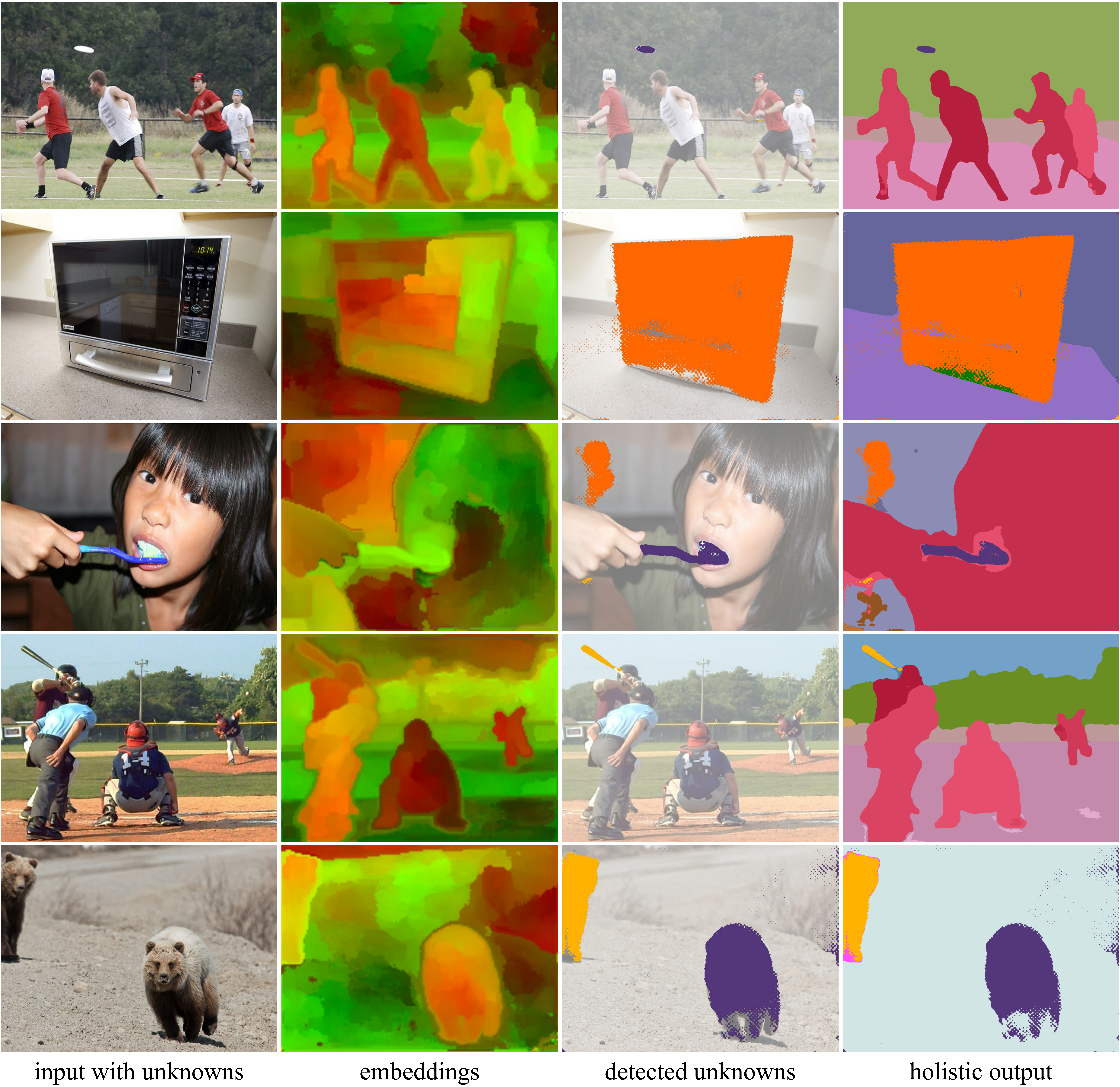}
\vspace{-0.7cm}
\end{center}
   \caption{Example predictions of the proposed \pname\ on OOD data containing the held out classes of MS COCO~\cite{lin2014coco}. Held out classes (unseen unknowns) in the samples: \textit{frisbee}, \textit{microwave}, \textit{toothbrush}, \textit{baseball bat}, and \textit{bear}. All samples are equally resized. Input, embeddings, detected unknowns, and holistic output are shown.}
\label{fig:coco_suppl}
\vspace{-0.2cm}
\end{figure*}
%\setfigurecounter{11}

\begin{figure*}[t]
\begin{center}
\includegraphics[width=1.00\textwidth]{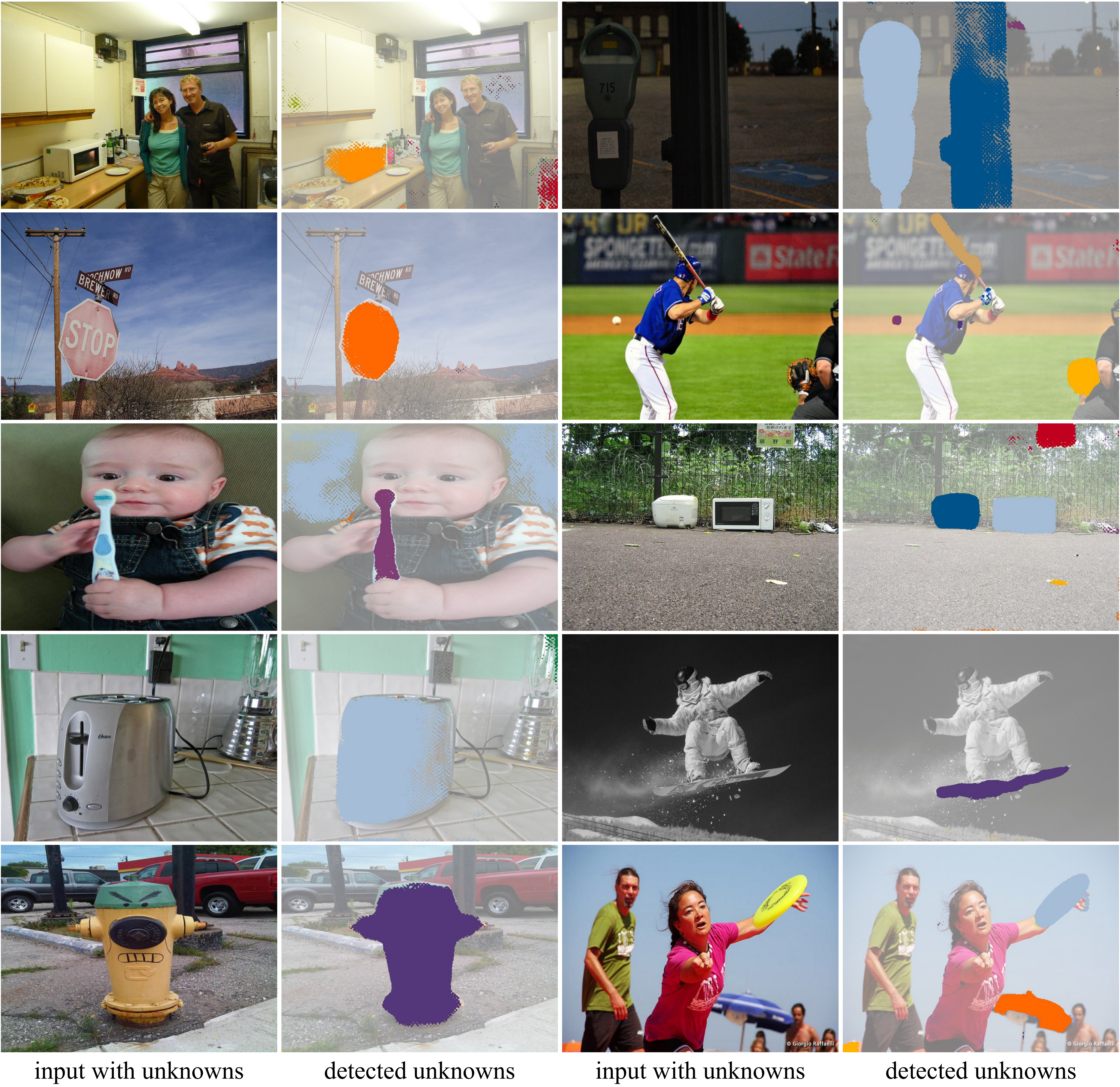}
\vspace{-0.7cm}
\end{center}
   \caption{Example predictions of the proposed \pname\ on OOD data containing the held out classes of MS COCO~\cite{lin2014coco}. Held out classes (unseen unknowns) in the samples: \textit{microwave}, \textit{parking meter}, \textit{stop sign}, \textit{baseball bat}, \textit{toothbrush}, \textit{toaster}, \textit{snowboard}, \textit{fire hydrant}, and \textit{frisbee}. Other unknown objects are included in the samples, such as the umbrella and the rice cooker (i.e., not part of the known classes). All samples are equally resized. Input and detected unknowns are shown.}
\label{fig:coco_suppl_simple}
\vspace{-0.2cm}
\end{figure*}
%\setfigurecounter{12}

\begin{figure*}[t]
\begin{center}
\includegraphics[width=1.00\textwidth]{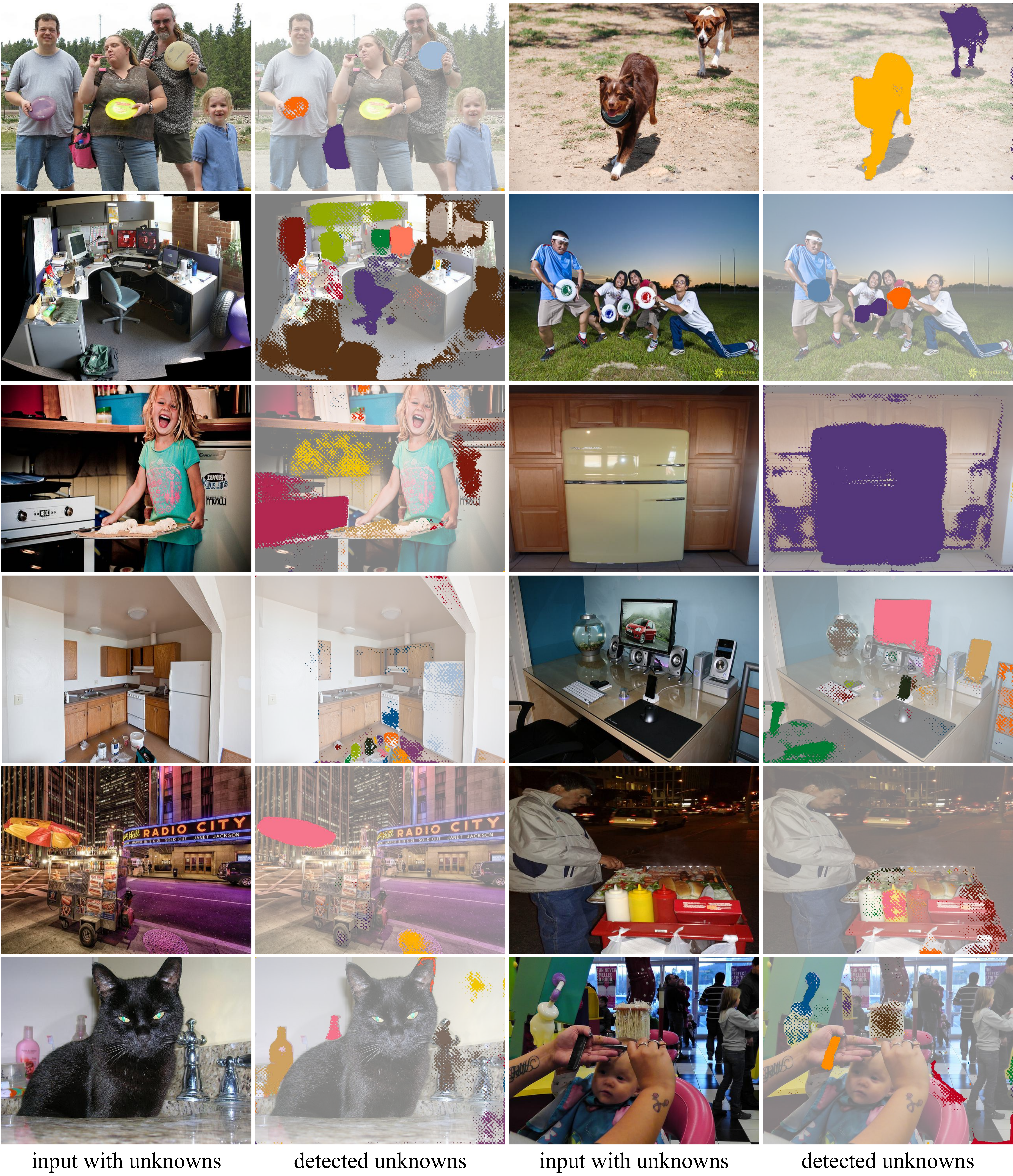}
\vspace{-0.7cm}
\end{center}
   \caption{Failure predictions of the proposed \pname\ on OOD data containing the held out classes of MS COCO~\cite{lin2014coco}. Held out classes (unseen unknowns) in the samples: \textit{frisbee}, \textit{keyboard}, \textit{mouse}, \textit{refrigerator}, \textit{hot dog}, \textit{toothbrush}, and \textit{scissors}. Other unknown objects are included in the samples, such as bag and comb (i.e., not part of the known classes). All samples are equally resized. Input and detected unknowns are shown.}
\label{fig:coco_failure}
\vspace{-0.2cm}
\end{figure*}
%\setfigurecounter{13}

\begin{figure*}[t]
\begin{center}
\includegraphics[width=1.00\textwidth]{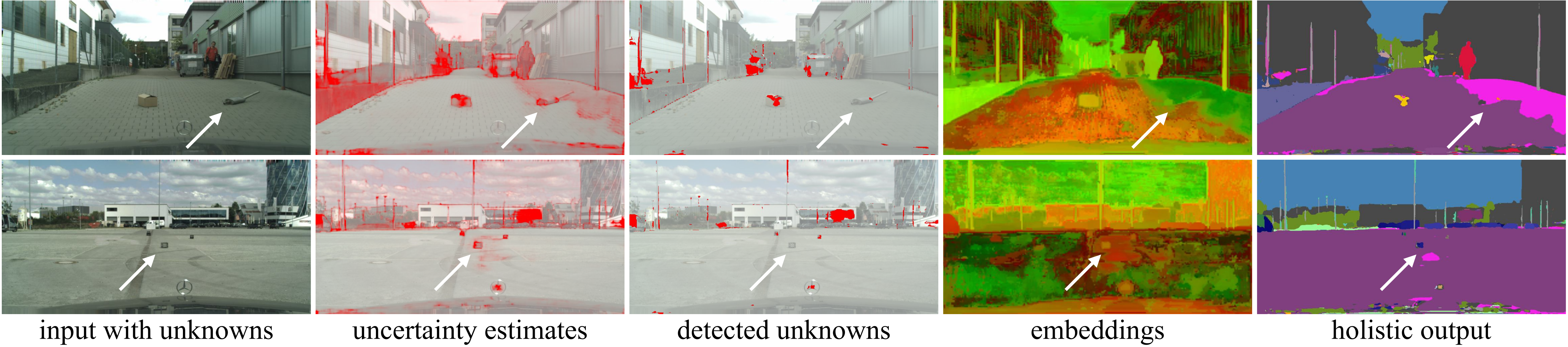}
\vspace{-0.7cm}
\end{center}
   \caption{Failure predictions of the proposed \pname\ on OOD data from the test set of Lost\&Found~\cite{pinggera2016lost}, with a transfer from Cityscapes~\cite{cordts2016cityscapes}. White arrows mark missed OOD objects as the estimated uncertainty was relatively low and filtered out.}
\label{fig:uncertainty_low}
\vspace{-0.2cm}
\end{figure*}
%\setfigurecounter{14}

% the benefit of uncertainty
Nevertheless, as shown already in Section~\reff{5}, estimating the uncertainty allows to properly cope with unknown objects by adding an extra layer of prediction. %~\ref{sec:exp_results}
Contrary to the idea of prior works (Section~\reff{2}) of predicting unknowns via the \textit{void} class, which directly competes with the other semantic classes for being part of the output, uncertainty estimates go on top of the standard semantic predictions. %~\ref{sec:rel_work}
Although this complicates dealing with multiple network outputs, it offers a wider spectrum and deeper insights since the uncertainty could be ignored or considered with various thresholds depending on the situation (Figure~\ref{fig:plot}), for the same trained model and output. Since estimating the uncertainty aims at smoothly quantifying the domain gap from the training data, we believe it is better suited to highly unpredictable unseen real-world scenarios as in holistic segmentation settings.

% the poor instancing
Furthermore, Figure~\ref{fig:qualitative_comp} shows the capability of each method to identify instances of unknown OOD objects. For both approaches, this is related to the clustering of embeddings corresponding to those pixels predicted as unknown, via \textit{void} (OSIS), or as highly uncertain (ours). In particular, OSIS tended to over-fragment unknown objects into several small instances, as seen in the fourth, sixth, eighth, and last images. This again proves our modifications' effectiveness when dealing with the embeddings, as described in Section~\reff{4} and evaluated in Table~\reff{3}. %~\ref{sec:framework} ~\ref{table:pano_fused}
Additionally, OSIS could not distinguish the two neighboring OOD objects in the sixth image.
% the big black
Moreover, OSIS often improperly assigned large regions to the same unknown instance. Similarly to ours, OSIS considers every unknown segment as part of an instance. By learning and predicting the \textit{void} class, during training, OSIS learned to precisely segment the bonnet of the ego car (labeled as \textit{void} in Cityscapes~\cite{cordts2016cityscapes}). However, at test time on Lost\&Found it could not tell the ego vehicle bonnet apart from a wide variety of pixels. This was the case for the unknown object in the seventh image, which was entirely assigned to the same instance as the bonnet or many other segments around knowns and unknowns (colored in black). The ego vehicle bonnet unknown instance (black) often surrounded other predicted unknown instances (e.g., in the second, fourth, sixth, eighth, and last images).

% unlabeled OOD objects
A benefit of estimating the uncertainty is the ability to account for a wide array of unusual regions. This is valuable for downstream tasks, e.g., trajectory prediction and path planning. Specifically, uncertainty estimates by the proposed \pname\ were high on the stroller in Figure~\reff{1}, as well as in Figure~\reff{4} on the walking assistance device on the left of the upper image and the cart pushed by the man waving on the right of the bottom image in Figure~\ref{fig:qualitative_suppl},
%and on the cart used by the man waving on the right of the fifth image,
none of which were labeled as unknown in the dataset~\cite{pinggera2016lost}, as they were not part of the objects manually placed by the authors. %~\ref{fig:teaser} ~\ref{fig:qualitative}
In Figure~\ref{fig:qualitative_comp}, this is repeated from a different perspective on the stroller in the background of the second image, the unusual van with the open doors in the fourth image, and the duffel bag in the sixth. By learning and predicting \textit{void}, OSIS ignored these unusual regions as it lacks the flexibility and granularity that our \pname\ offers by estimating the uncertainty.

\subsubsection{Additional Results on Unknowns}
\textbf{Lost\&Found}
Figure~\ref{fig:qualitative_suppl} shows additional qualitative outputs.
Once again, it can be seen how challenging the proposed holistic segmentation setting is. As in the predictions of Figure~\reff{4}, the model can distinguish most unknown objects. %~\ref{fig:qualitative}
It can be seen how specific areas of the images trigger higher uncertainty estimates. This is the case of the fences in the second and third images of Figure~\ref{fig:qualitative_suppl}, as well as unknown objects not part of the OOD labels of Lost\&Found, such as the cart on the right of the bottom image, as previously mentioned. As previously seen, \textit{stuff} structures (e.g., fences) are assigned to a single coherent instance ID throughout the whole image. At the same time, unusual objects (e.g., the cart in the last picture) have their dedicated ID. Figure~\ref{fig:qualitative_suppl} also provides some examples of unusual scenes present in the Lost\&Found~\cite{pinggera2016lost} dataset, posing significant challenges compared to Cityscapes~\cite{cordts2016cityscapes}.

\textbf{MS COCO}
Figures~\ref{fig:coco_suppl} and~\ref{fig:coco_suppl_simple} show qualitative outputs of the proposed \pname\ on the held out classes of MS COCO~\cite{lin2014coco}. The images report the vast diversity of the dataset, ranging from outdoor scenes to indoor close-ups. Remarkably, \pname\ delivered precise segments for unknown objects, correctly segmenting instances of individual unknowns despite their similarity with other objects of the same type in the same input, with reasonable estimations of known classes too (Figure~\ref{fig:coco_suppl}). Due to the difficulty of this problem, only a handful of segments are perfect, leaving room for improvements for known and unknown objects. Thanks to its strong uncertainty estimation capabilities, the proposed \pname\ not only identified the held-out classes but also other unknown objects which are not part of the set of known classes, such as the rice cooker and the umbrella on the right of Figure~\ref{fig:coco_suppl_simple} in the third and bottom rows respectively. This shows the efficacy of our method on a wide variety of scenarios typical of the real world.

\subsubsection{Failure Cases}
%\textbf{Considerations on estimating the uncertainty}
\textbf{MS COCO}
While the proposed \pname\ delivers reasonable estimates in various settings, from indoor to outdoor, the problem at hand is highly challenging, and its predictions are not perfect, as confirmed by the quantitative results. 
Figure~\ref{fig:coco_failure} reports failure cases on a set of challenging samples of MS COCO.
It can be seen that while \pname\ identified unknowns reasonably, it often missed those objects that are part of the evaluated held-out categories. A series of issues cause this. In some instances, the integrated uncertainty estimates could not fully discover the unseen unknowns, e.g., only partially detected for both \textit{refrigerators} at the left of the third and fourth rows. We also noticed systematic issues with certain classes, such as \textit{keyboard}, \textit{mouse} (both in the right of the fourth row), \textit{hot dog} (in the fifth row), and \textit{scissors} (right of the last row). This could be attributed to these objects being semantically relatively close to known classes, such as \textit{sandwich} for \textit{hot dog}. Small objects were hard to see and therefore ignored by our method. This is the case of the \textit{toothbrush} behind the \textit{cat} in the last row of the figure. \pname\ also had difficulties telling apart from one another very close and similar unknowns, e.g., the central \textit{frisbees} in the right of the second row differing for the logo's color. However, it could successfully separate neighboring objects on multiple occasions, such as the containers on the ground of the left image in the fourth row (not evaluated in this experiment). Moreover, we noticed difficulties with particularly unusual inputs and cluttered environments, e.g., on the left of the second row. With other inputs, the uncertainty of \pname\ was also triggered on known objects, such as the \textit{dogs} on the top right (albeit correctly separated into two instances), or the cabinets on the right of the third row.
Since several held-out classes contained everyday kitchen-related items (e.g., \textit{refrigerator}, \textit{microwave}, and \textit{toaster}), or typical desk objects (e.g., \textit{keyboard} and \textit{mouse}), the model could see only a handful of kitchens and offices (i.e., those images where none of these held out objects appeared), which severely impacted its ability to handle these situations appropriately. This is a limitation of holding out samples from MS COCO, compared to using a dedicated separate dataset, such as Lost\&Found~\cite{pinggera2016lost}.
Furthermore, given that \pname\ estimates the model uncertainty, more training data covering a wider variety of scenarios could be beneficial to further reduce the uncertainty on the known classes and improve the known-unknown boundary of \pname.

\textbf{Lost\&Found}
Figure~\ref{fig:uncertainty_low} shows failure cases caused by the necessary filtering of the uncertainty estimates. While the uncertainty was triggered by a variety of unusual areas, including the vast majority of unknown objects, its a priori filtering (based on closed-set training data, Section~\reff{4.2}) sometimes caused the unknown object to be completely undetected. %~\ref{sec:open_pano}
Although this filtering is aimed at removing low uncertainty areas which are probably in-domain (e.g., the fence in the upper image), it could 
inadvertently remove proper OOD objects (e.g., those marked by the white arrows). This is related to the trade-off shown in Figure~\ref{fig:plot}, so keeping more unknowns (i.e., lower threshold $t$) reduces the in-domain performance. Nevertheless, in the embeddings visualizations, the model correctly isolated the entire marked box in the lower image and precisely segmented the cardboard box in the upper one. However, the two unknown objects were not detected, due to the difficulty of merging multiple outputs and interpreting uncertainty estimates without access to OOD data. It should be considered that the proposed \pname\ does not distinguish between the uncertainty for unknown objects and that of unusual known classes. The difference might lie in the amount of uncertainty corresponding to these regions, hence the filtering via the threshold $t$ to attempt telling apart completely unknown from unusual, which remains highly challenging without using any information about unknowns at training time, as in our setup.
%\input{appendix/appendix.tex}

%%%%%%%%% REFERENCES
{\small
\bibliographystyle{ieee_fullname}
\bibliography{egbib}
}

%\clearpage
%\input{appendix/appendix.tex}

\end{document}